\documentclass[letterpaper]{article} 
\usepackage{aaai2026}  
\usepackage{times}  
\usepackage{helvet}  
\usepackage{courier}  
\usepackage[hyphens]{url}  
\usepackage{graphicx} 
\usepackage{natbib} 
\usepackage{caption}
\usepackage{booktabs} 
\usepackage{multirow} 
\usepackage{amssymb}
\usepackage{pifont}
\usepackage{algorithm}
\usepackage{algorithmic}
\usepackage{newfloat}
\usepackage{listings}
\DeclareCaptionStyle{ruled}{labelfont=normalfont,labelsep=colon,strut=off} 
\floatstyle{ruled}
\newfloat{listing}{tb}{lst}{}
\floatname{listing}{Listing}

\title{ALScope: A Unified Toolkit for Deep Active Learning}

\nocopyright
\author {
    Chenkai Wu\textsuperscript{\rm 1},
    Yuanyuan Qi\textsuperscript{\rm 1},
    Xiaohao Yang\textsuperscript{\rm 1},
    Jueqing Lu\textsuperscript{\rm 1},
    Gang Liu\textsuperscript{\rm 2},
    Wray Buntine\textsuperscript{\rm 3},
    Lan Du\textsuperscript{\rm 1}\thanks{Corresponding author.}
}
\affiliations {
    \textsuperscript{\rm 1}Monash University, Australia\\
    \textsuperscript{\rm 2}Harbin Engineering University, China\\
    \textsuperscript{\rm 3}VinUniversity, Vietnam\\
    cwuu0040@student.monash.edu, \{Yuanyuan.Qi, Xiaohao.Yang, Jueqing.Lu, Lan.Du\}@monash.edu, \\liugang@hrbeu.edu.cn, wray.b@vinuni.edu.vn
}

\usepackage{booktabs}
\usepackage{array}
\usepackage{amsmath}

\begin{document}

\maketitle

\begin{abstract}
Deep Active Learning (DAL) reduces annotation costs by selecting the most informative unlabeled samples during training. As real-world applications become more complex, challenges stemming from distribution shifts (e.g., open-set recognition) and data imbalance have gained increasing attention, prompting the development of numerous DAL algorithms. However, the lack of a unified platform has hindered fair and systematic evaluation under diverse conditions. Therefore, we present a new DAL platform ALScope for classification tasks, integrating 10 datasets from computer vision (CV) and natural language processing (NLP), and 21 representative DAL algorithms, including both classical baselines and recent approaches designed to handle challenges such as distribution shifts and data imbalance. This platform supports flexible configuration of key experimental factors, ranging from algorithm and dataset choices to task-specific factors like out-of-distribution (OOD) sample ratio, and class imbalance ratio, enabling comprehensive and realistic evaluation. We conduct extensive experiments on this platform under various settings. Our findings show that: (1) DAL algorithms' performance varies significantly across domains and task settings; (2) in non-standard scenarios such as imbalanced and open-set settings, DAL algorithms show room for improvement and require further investigation; and (3) some algorithms achieve good performance, but require significantly longer selection time. 

\end{abstract}

\begin{links}
     \link{Code}{https://github.com/WuXixiong/DALBenchmark.}
\end{links}

\section{Introduction}

Deep learning models have demonstrated exceptional performance across various fields. 
However, training such models typically demands large amounts of labeled data, resulting in high annotation costs, especially in domains like medical \cite{chen2024think, zheng2023evidential, hoi2006batch}, finance \cite{smailovic2014stream, farquad2012analytical}, or chemistry \cite{warmuth2001active}, where data annotation requires human expertise. Active Learning (AL) \cite{zhan2022comparative, ren2021survey, li2024survey} is a well-established approach to reducing annotation costs by strategically selecting the most informative data for labeling, enabling a reduction in labeled data requirements while preserving model performance. In recent years, the combination of deep learning and active learning—known as Deep Active Learning (DAL)—has advanced rapidly, giving rise to numerous new methods and their gradual adoption across various domains, including image classification \cite{ranganathan2017deep, jin2022deep, li2024unlabeled}, semantic segmentation \cite{xie2023annotator, hwang2023active}, object detection \cite{park2023active, yang2024plug, hekimoglu2024monocular, greer2024language, yamani2024active}, and more.

\begin{table*}[ht]
\centering
\begin{tabular}{l c c c c c}
\toprule
\textbf{Platform} & \textbf{\# DAL algorithms} & \textbf{CV} & \textbf{NLP} & \textbf{Open-set} & \textbf{Imbalance-set} \\
\midrule
DeepAL+ \cite{zhan2022comparative}    & 15 & \checkmark & \ding{55} & \ding{55} & \ding{55} \\
Baal \cite{atighehchian2020bayesian}        & 8  & \checkmark & \ding{55} & \ding{55} & \ding{55} \\
Eval-AL \cite{ji2023randomness}     & 7  & \checkmark & \ding{55} & \ding{55} & \ding{55} \\
CDALBench \cite{werner2024cross}  & 12 & \checkmark & \checkmark & \ding{55} & \ding{55} \\
Ours        & 21 & \checkmark & \checkmark & \checkmark & \checkmark \\
\bottomrule
\end{tabular}
\caption{Comparison of different DAL platforms.}
\label{tab:dal-benchmark-comparison}
\end{table*}

As more DAL algorithms emerge, the lack of standardized experimental setups and the proliferation of diverse codebases have made comprehensive performance benchmarking increasingly challenging. 
In particular, new algorithms are often compared only against traditional baselines, 
leading to ambiguous performance comparisons among recent methods. 
Critical factors such as initial labeled sample size and the query batch size, both of which significantly impact performance, are inconsistently configured across studies, resulting in divergent and sometimes contradictory findings.
Additionally, the use of disparate codebases across works imposes a substantial burden on researchers attempting to benchmark new algorithms against existing ones.
Furthermore, there is growing interest in evaluating DAL algorithms
under real-world conditions involving distribution shift between labeled and unlabeled samples \cite{bengar2022class, ning2022active, safaei2024entropic, mao2024inconsistency}.
These challenges highlight the urgent need for a unified DAL platform that
enables consistent, reproducible, and flexible evaluation—particularly in scenarios reflective of practical deployment.

In this paper, we present a novel DAL platform ALScope that unifies experimental setups and standardizes implementations for evaluating DAL algorithms. 
Our platform supports 21 well-established DAL algorithms, 
spanning a comprehensive range of approaches.
These include uncertainty-based algorithms (\cite{lewis1995sequential}, \cite{shannon2001mathematical}, \cite{scheffer2001active}, \cite{kampffmeyer2016semantic}, \cite{gal2017deep}, \cite{freeman1965elementary}, \cite{yoo2019learning}), diversity-based algorithm (\cite{sener2017active}), hybrid algorithms (\cite{parvaneh2022active}, \cite{ash2019deep}, \cite{tan2023bayesian}, \cite{tan2023bayesian}, \cite{kim2023saal}, \cite{li2024deep}, \cite{kye2023TiDAL}), algorithms for handling imbalanced datasets (\cite{bengar2022class}, \cite{bengar2022class}), algorithms for handling open-set data (\cite{du2021contrastive}, \cite{ning2022active}, \cite{park2022meta}). 
Focusing on classification tasks, 
our platform includes widely used datasets such as AGNEWS \cite{zhang2015character}, YELP \cite{yelp_dataset_challenge}, SST-5 \cite{socher2013recursive}, TREC-6 \cite{li2002learning}, CIFAR-10/100 \cite{krizhevsky2009learning}, and TinyImageNet \cite{le2015tiny}. 
It further enables real-world evaluation by providing options to convert these datasets into open-set or imbalanced versions.
ALScope offers flexibility setting key experimental parameters, including backbone architecture, initial sample size, query size, OOD rate, and dataset imbalance rate.
Its modular design also facilitates the seamless integration of new algorithms, 
supporting reproducibility and future innovation in DAL research.

ALScope introduces several key advancements over existing DAL platforms \cite{zhan2022comparative, atighehchian2020bayesian, ji2023randomness, werner2024cross}, in the following aspects:
\begin{itemize}
    \item \textbf{Algorithm coverage}: Most existing DAL platforms include only a limited number of the latest algorithms. To address this limitation, our platform incorporates a comprehensive collection of state-of-the-art methods, reflecting the latest developments in the field. 
    \item \textbf{Domain Diversity}: Existing platforms  \cite{zhan2022comparative, ji2023randomness} predominantly focus on a single domain, such as CV, and lack coverage of other task types. Addressing this gap, our platform encompasses both CV and NLP, providing comprehensive support for classification tasks in both domains. This significantly enhances the platform's applicability and generality.
    \item \textbf{Realistic evaluation settings}: Existing platforms primarily evaluate DAL under closed-set conditions. Our platform extends this by supporting more realistic scenarios, such as open-set and class-imbalanced setting, which better reflect the challenges encountered in real-world applications.
\end{itemize}

Using the proposed platform, we conducted systematic comparisons across multiple tasks and DAL algorithms, examining key factors that may influence performance, such as query batch size and different task settings including open-set and imbalanced conditions. 
This evaluation framework enables fair and consistent assessment of DAL algorithms
under different conditions, reveals the strengths and weaknesses of different algorithms, 
and offers insights for future developments.

\section{Related Work}
In this section, we provide a brief overview of the principal categories of DAL algorithms that are widely explored in the literature. For a more comprehensive discussion, we refer the reader to existing survey articles on active learning \cite{ren2021survey, zhan2022comparative, li2024survey}. In addition, we provide a brief discussion of existing benchmark platforms for active learning.

\subsection{DAL Algorithms}
\textbf{Uncertainty-based} sampling methods \cite{shannon2001mathematical, wang2014new, netzer2011reading, kampffmeyer2016semantic, gal2017deep, smith2023prediction, kye2023TiDAL} aim to select the samples for which the model is least confident, such as those near the decision boundary. These methods are effective in identifying informative samples that significantly improve the model with fewer labeled instances.

\textbf{Diversity-based} sampling selects a representative subset that covers the overall data distribution \cite{sener2017active, gissin2019discriminative, sinha2019variational}. These methods aim to avoid redundancy and maximize the diversity of queried samples using techniques like core-set selection or clustering \cite{sener2017active}.

\textbf{Hybrid} approaches combine uncertainty and diversity strategies to balance informativeness and representativeness \cite{shui2020deep, ash2019deep, parvaneh2022active, li2024deep}. For example, BADGE \cite{ash2019deep} uses gradient embeddings to capture both uncertainty and diversity in batch selection.

\subsection{DAL Under Real-World Scenarios}
\paragraph{DAL with Class Imbalance} addresses the common real-world scenario in which class distributions are inherently imbalanced \cite{bengar2022class, lesci2024anchoral, aggarwal2021minority, nuggehalli2023direct}. For example, CBAL \cite{bengar2022class} addresses this by modifying the acquisition function to incorporate class-frequency compensation, improving performance on underrepresented classes.

\paragraph{Open-Set DAL} relaxes the closed-set assumption by allowing the unlabeled data to contain instances from classes that are not present in the labeled set \cite{ning2022active, park2022meta, yang2024not, safaei2024entropic, mao2024inconsistency}. For example,  \citet{ning2022active} handles this by filtering out unknown-class instances using a threshold over the maximum activation value (MAV), combined with uncertainty sampling.

\subsection{Existing Platforms}

Existing DAL platforms have certain limitations in different aspects. 
DeepAL+ only supports algorithms released up to 2020 and has yet to incorporate the latest 
methods. 
Baal, with its focus on Bayesian-based active learning algorithms, has a narrower scope for broader comparisons across different algorithms. The Eval-al framework currently 
includes only seven classic DAL algorithms and primarily uses the CIFAR-10 and CIFAR-100 datasets. 
Moreover, existing platforms primarily focus on standard active learning settings within computer vision tasks.
Those limitations restrict the platform’s capacity to support broader and more in-depth experimental evaluations.
In contrast, ALScope offers several advantages over existing platforms:
(1) It supports a wide range of state-of-the-art and recent AL algorithms;
(2) It includes diverse datasets spanning both computer vision and natural language processing tasks;
(3) It goes beyond standard AL settings by incorporating more realistic scenarios such as class imbalance and open-set active learning.

\begin{figure*}[htbp]
    \centering
    \includegraphics[width = 0.8\textwidth]{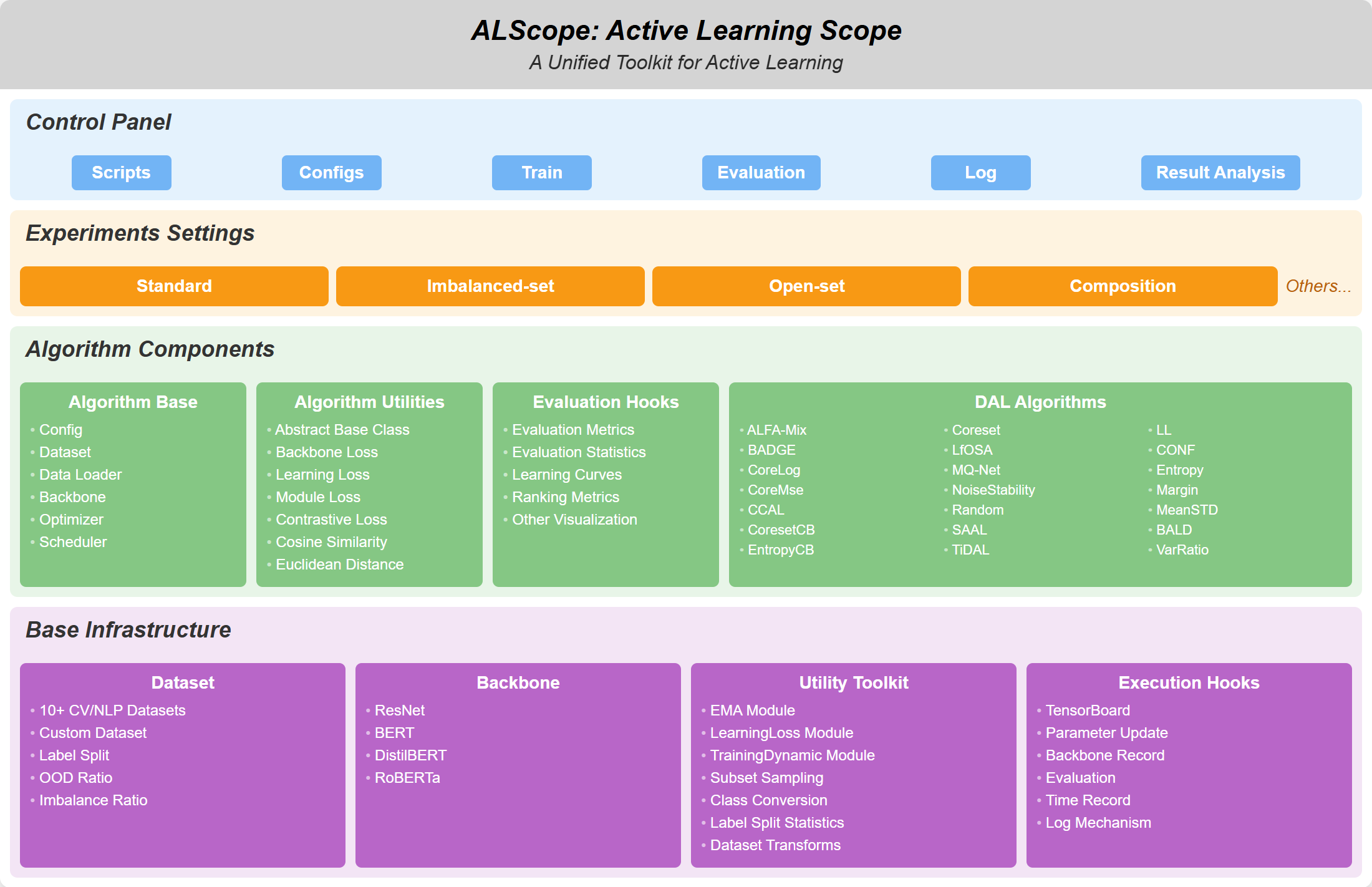}
    \caption{High-level platform framework of ALScope.}
    \label{fig:dalb}
\end{figure*}

\section{DAL Algorithms in ALScope}
\label{al_algorithms}
We implement 21 DAL algorithms in the ALScope codebase, including uncertainty-based algorithms: CONF \cite{lewis1995sequential}, Entropy \cite{shannon2001mathematical}, Margin \cite{scheffer2001active}, MeanSTD \cite{kampffmeyer2016semantic}, BALD \cite{gal2017deep}, VarRatio \cite{freeman1965elementary}, LL \cite{yoo2019learning}; diversity-based algorithm: Coreset \cite{sener2017active}; hybrid algorithms: ALFA-Mix \cite{parvaneh2022active}, BADGE \cite{ash2019deep}, CoreLog \cite{tan2023bayesian}, CoreMSE \cite{tan2023bayesian}, SAAL \cite{kim2023saal}, NoiseStability \cite{li2024deep}, TiDAL \cite{kye2023TiDAL};, algorithms for handling imbalanced datasets: CoresetCB \cite{bengar2022class}, EntropyCB \cite{bengar2022class}; algorithms for handling open-set data: CCAL \cite{du2021contrastive}, LfOSA \cite{ning2022active}, MQ-Net \cite{park2022meta}; and Random Sampling.

The implementation of all DAL algorithms is primarily based on the official open-source code released in the corresponding publications, in order to ensure maximum reproducibility of experimental results and accuracy of method implementation, further details of these algorithms can be found in the Appendix. 

\begin{table}[ht]
\centering
\begin{tabular}{lccc}
\toprule
\textbf{Dataset} & \textbf{\#Training data} & \textbf{\#Test data} & \textbf{\#Class} \\
\midrule
CIFAR-10            & 50,000          & 10,000  & 10 \\
CIFAR-100           & 50,000          & 10,000  & 100 \\
TinyImageNet        & 100,000         & 10,000  & 200 \\
SVHN                & 73,257          & 26,032  & 10 \\
MNIST               & 60,000          & 10,000  & 10 \\
\midrule
YelpReview          & 700,000         & 140,000 & 5 \\
AG News             & 100,000         & 7,600   & 4 \\
SST-5               & 11,512          & 1,152   & 5 \\
TREC-6              & 6,000           & 700     & 6 \\
IMDB                & 25,000          & 25,000  & 2 \\
\bottomrule
\end{tabular}
\caption{Dataset Statistics in ALScope}
\label{table:dataset}
\end{table}

\section{Tasks and Datasets}

ALScope includes 10 datasets commonly used in the computer vision and natural language processing domains, respectively. Each dataset can be utilized in standard DAL settings—namely, closed-set and class-balanced scenarios—as well as in more realistic settings such as open-set recognition and class imbalance, which reflect challenges observed in real-world applications.

Each dataset included in ALScope satisfies the following criteria, justifying its selection: (1) It is a commonly used benchmark in DAL experiments. (2) It is widely adopted across a range of domains, reflecting its popularity and general applicability. (3) It has a moderate scale in terms of size, making it computationally feasible and accessible for researchers. A summary of the dataset statistics is provided in Table~\ref{table:dataset}.

\subsection{CV Tasks}  
The CV datasets included in ALScope are as follows: (1) \textit{CIFAR-10}~\cite{krizhevsky2009learning}: 60,000 32$\times$32 color images across 10 classes (e.g., airplane, dog, truck); (2) \textit{CIFAR-100}~\cite{krizhevsky2009learning}: same number of images as CIFAR-10 but divided into 100 finer-grained classes; (3) \textit{TinyImageNet}~\cite{Le2015TinyIV}: a 64$\times$64 subset of ImageNet \cite{deng2009imagenet} with 200 classes and 500 training images per class, exhibiting greater intra-class variance; (4) \textit{SVHN}~\cite{netzer2011reading}: real-world digit images (0--9) from Google Street View, with 10 classes; and (5) \textit{MNIST}~\cite{lecun1998gradient}: 28$\times$28 grayscale handwritten digit images over 10 classes (0--9). CIFAR-10 and CIFAR-100 are extensively used in deep active learning, while TinyImageNet introduces higher complexity. SVHN and MNIST further broaden the benchmark’s coverage of diverse classification tasks.

\subsection{NLP Tasks}  
The NLP datasets included in ALScope cover a wide range of text classification tasks and are widely used in active learning research. These include: (1) \textit{AG News}~\cite{zhang2015character}: 4 news topic classes (World, Sports, Business, Sci/Tech); (2) \textit{Yelp Review}~\cite{yelp_dataset_challenge}: 5-class sentiment analysis; (3) \textit{IMDB}~\cite{maas2011learning}: binary sentiment classification (positive/negative); (4) \textit{SST-5}~\cite{socher2013recursive}: 5 sentiment classes for fine-grained movie review analysis; (5) \textit{TREC-6}~\cite{li2002learning}: 6 question types (e.g., entity, location, number); and (6) \textit{DBpedia}~\cite{lehmann2015dbpedia}: 14 ontology-based classes for knowledge base entry categorization. These datasets differ in the number of classes, text length, and linguistic complexity, enabling a thorough evaluation of the generalization and robustness of active learning methods across diverse classification contexts.

\subsection{From Standard to Real-World Scenarios}
\label{real_world}
Standard AL experimental settings assume idealized conditions, where datasets are class-balanced and follow a closed-set assumption. However, in real-world scenarios, data distributions are often imbalanced across classes, and the closed-set assumption may not hold. For example, models may encounter previously unseen classes at test time—commonly referred to as the open-set setting. These challenges are important to consider, as they significantly affect the design and evaluation of AL algorithms in practical applications.

ALScope supports a flexible construction of class-imbalanced and/or open-set settings for each dataset, enabling more realistic simulation of real-world learning conditions. The transformation from standard AL settings to open-set or class-imbalanced configurations is governed by two key factors: 
\paragraph{(1) OOD Ratio} determines the proportion of OOD data introduced into the unlabeled dataset, 
which is defined as:
\begin{equation}
\text{OOD Ratio} = \frac{N_{\text{OOD}}}{N_{\text{unlabeled}}},
\end{equation}
where \( N_{\text{OOD}} \) is the number of OOD samples and \( N_{\text{unlabeled}} \) is the total number of samples in the unlabeled dataset. 
\paragraph{(2) Imbalance Ratio} controls the degree of skewness in the class distribution, often following a long-tail distribution. The number of samples for each class \( i \) is computed as:
\begin{equation}
n_i = n_{\text{max}} \cdot \gamma^{\frac{i}{C - 1}}, \quad i = 0, 1, \ldots, C-1,
\end{equation}
where \( n_i \) is the number of samples in class \( i \), \( n_{\text{max}} \) is the number of samples in the most frequent class, \( \gamma \in (0, 1] \) is the ratio between the number of samples in the least frequent class and the number of samples in the most frequent class in the dataset, and \( C \) is the total number of classes.

\section{Platform Structure}
In this section, we provide an overview of the structure of ALScope. Our toolkit adopts a modular design, with the overall architecture divided into four functional modules: Control Panel Module, Experiments Settings Module, Algorithm Components Module, and Base Infrastructure Module, as shown in Figure~\ref{fig:dalb}. Each module features a clear structure and well-defined responsibilities, facilitating efficient development and experimental management for active learning algorithms.

\paragraph{Control Panel Module}
This module streamlines the DAL workflow through modular components and intuitive command-line interfaces (see the Appendix for example scripts, commands, configs, and logs), and includes the following key functionalities:  
\textbf{(A) Scripts}: Predefined and customizable scripts support automation of training and evaluation, and can be integrated into batch pipelines for large-scale experiments.  
\textbf{(B) Configs}: Flexible configuration files enable quick setup of datasets, models, strategies, and metrics, promoting reproducibility.  
\textbf{(C) Train}: A simple command triggers training with automatic management of iterative cycles for improved efficiency.  
\textbf{(D) Evaluation}: Model evaluation is available at any stage and can be embedded in automated workflows.  
\textbf{(E) Log}: Comprehensive logs capture progress, metrics, and model behavior for effective tracking and analysis.  
\textbf{(F) Result Analysis}: Post-processing and visualization tools facilitate interpretation and comparison of results across algorithms and settings.

\paragraph{Experiments Settings Module}
This module enables flexible configuration of DAL experiments across diverse real-world scenarios, supporting a wide range of needs from standard setups to complex extensions. It currently supports the following settings:
\textbf{(1) Standard Setting}:
The default active learning workflow, serving as a baseline configuration.
\textbf{(2) Imbalance Setting}:
Supports the construction of class-imbalanced learning scenarios, allowing evaluation of algorithm performance under skewed data distributions.
\textbf{(3) Open-set Setting}:
Enables open-set active learning experiments, where models are required to identify and manage previously unseen classes during the learning process.
\textbf{(4) Compositional Setting}:
Allows the integration of multiple settings—such as imbalanced and open-set scenarios—to simulate more complex and realistic experimental environments.

This module also facilitates flexible ablation study design, enabling assessment of the impact of individual algorithmic components on overall performance. In addition, it offers fine-grained control over key parameters—such as the size of the initial labeled set, unlabeled pool, query batch, and the number of active learning cycles—allowing precise tuning and realistic scenario simulation.

\paragraph{Algorithm Components Module}

In this module, we have implemented abstract base classes for various AL algorithms. This abstraction significantly enhances code reusability and facilitates the integration of new AL algorithms into ALScope by researchers. In addition to the algorithm-specific computational components, the module provides a rich set of shared \textbf{Algorithm Utilities}, enabling researchers to leverage these reusable tools in the construction of novel algorithms. Furthermore, we offer comprehensive \textbf{evaluation hooks} to assess the performance of each algorithm. These include a variety of evaluation metrics such as accuracy and F1-score, as well as experimental statistics, learning curves, ranking matrices, and other visualizations—allowing for multi-dimensional evaluation of AL algorithms. Building upon this infrastructure, ALScope currently supports 21 AL algorithms. 

\paragraph{Base Infrastructure Module}

In this module, we have implemented essential core functionalities required for training AL algorithms. This module also includes the implementation of dataset loaders, data processing pipelines, and backbone models used throughout ALScope. To enable a flexible and modular training process, we have defined customized training functions for various AL algorithms, while also integrating a shared \textbf{Utility Toolkit} that provides commonly used components applicable across multiple algorithms. Additionally, we have incorporated \textbf{Execution Hooks} to support comprehensive experiment monitoring and debugging by recording training time, model states, evaluation results, and log information.

\begin{table*}[htbp]
\centering
\setlength{\tabcolsep}{1mm}
\footnotesize
    \begin{tabular}{llccc|ccc|ccc|ccc}
    \toprule
    \multirow{2}{*}{} & \bfseries Dataset & \multicolumn{3}{c}{CIFAR-10} & \multicolumn{3}{c}{CIFAR-100} & \multicolumn{3}{c}{SST-5} & \multicolumn{3}{c}{YELP-3000} \\
    \cmidrule{3-14}
    & \bfseries Setting & NORMAL & OOD & Imb & NORMAL & OOD & Imb & NORMAL & OOD & Imb & NORMAL & OOD & Imb \\
    \midrule
    \multirow{7}{*}{\bfseries Hybrid} & ALFA-Mix & 0.810 & 0.499 & 0.487 & 0.561 & 0.416 & 0.396 & 0.339 & 0.629 & \underline{0.363} & 0.405 & 0.674 & 0.445 \\
    & BADGE & 0.827 & 0.569 & \underline{0.503} & \textbf{0.578} & * & * & 0.316 & 0.630 & 0.356 & 0.390 & \underline{0.691} & 0.463 \\
    & SAAL & 0.802 & 0.534 & 0.261 & 0.560 & 0.414 & 0.229 & 0.273 & \underline{0.657} & 0.355 & 0.392 & 0.650 & 0.446 \\
    & TiDAL & 0.809 & 0.551 & \textbf{0.505} & 0.558 & 0.413 & 0.402 & - & - & - & - & - & - \\
    & CoreLog & 0.810 & 0.500 & 0.496 & 0.559 & 0.394 & 0.382 & 0.336 & 0.617 & 0.359 & 0.375 & \textbf{0.695} & 0.444 \\
    & CoreMSE & 0.807 & 0.541 & 0.497 & 0.557 & 0.409 & 0.407 & 0.335 & 0.636 & 0.341 & 0.382 & 0.639 & 0.442 \\
    & NoiseStability & 0.830 & 0.544 & 0.484 & \underline{0.575} & * & * & - & - & - & - & - & - \\
    \midrule
    \multirow{7}{*}{\bfseries Uncertainty} & BALD & 0.816 & 0.554 & 0.499 & 0.567 & 0.392 & 0.392 & 0.304 & 0.599 & \textbf{0.365} & 0.344 & 0.614 & 0.427 \\
    & CONF & \textbf{0.833} & 0.549 & 0.483 & 0.566 & 0.378 & 0.385 & \underline{0.340} & 0.651 & 0.356 & \underline{0.407} & 0.635 & 0.448 \\
    & Entropy & 0.824 & 0.550 & 0.490 & 0.559 & 0.365 & 0.382 & 0.321 & 0.651 & 0.356 & 0.370 & 0.635 & 0.418 \\
    & LL & 0.794 & 0.504 & 0.467 & 0.538 & 0.370 & 0.366 & - & - & - & - & - & - \\
    & Margin & \underline{0.832} & \underline{0.565} & 0.485 & 0.566 & 0.397 & 0.379 & 0.326 & 0.651 & 0.359 & 0.401 & 0.635 & 0.427 \\
    & MeanSTD & 0.809 & 0.553 & 0.307 & 0.557 & \underline{0.417} & 0.197 & \textbf{0.352} & 0.617 & 0.346 & 0.349 & 0.653 & \textbf{0.460} \\
    & VarRatio & \textbf{0.833} & 0.549 & 0.483 & 0.568 & 0.378 & 0.385 & \underline{0.340} & 0.651 & 0.356 & \underline{0.407} & 0.635 & 0.448 \\
    \midrule
    \bfseries Diversity & Coreset & 0.823 & 0.529 & 0.499 & 0.572 & 0.393 & \underline{0.409} & 0.323 & \textbf{0.658} & 0.350 & 0.385 & 0.641 & 0.452 \\
    \midrule
    \bfseries Open-set & LfOSA & 0.634 & 0.522 & 0.446 & 0.468 & \textbf{0.420} & 0.399 & 0.317 & 0.590 & 0.350 & 0.321 & 0.573 & 0.409 \\
    \midrule
    \multirow{2}{*}{\bfseries Imbalanced-set} & CoresetCB & 0.822 & 0.518 & 0.499 & 0.570 & 0.390 & \textbf{0.412} & 0.323 & \textbf{0.658} & 0.344 & \textbf{0.408} & 0.641 & \underline{0.457} \\
    & EntropyCB & 0.826 & 0.557 & 0.495 & 0.568 & * & * & 0.318 & 0.641 & 0.328 & 0.387 & 0.574 & 0.435 \\
    \midrule
     & Random & 0.809 & \textbf{0.570} & 0.481 & 0.562 & 0.407 & 0.392 & 0.324 & 0.647 & 0.356 & 0.381 & 0.686 & 0.461 \\
    \bottomrule
\end{tabular}
\caption{
Final accuracy of different DAL algorithms across various settings and datasets. Top two performers in each metric are highlighted: First place in bold, second place underlined. The numbers in the table represent the accuracy of the final training round. `*' indicates that the algorithm did not complete within a reasonable time under this experimental setting and therefore failed to produce a result. Please refer to Section~\ref{sec:time} and the Appendix for a more detailed discussion on runtime performance; `-' indicates that the algorithm currently does not support experiments on text datasets.
}
\label{table:main_results}
\end{table*}

\section{Platform Results}

We evaluate 19 DAL algorithms without additional training modules on seven datasets, including CIFAR-10, CIFAR-100, SST-5, and YELP-3000 (a 3,000-instance/class subset of YelpReviews). Each dataset is tested under three settings: \textbf{NORMAL}, \textbf{OOD} (60\% OOD samples), and \textbf{Imb} (class imbalance $\gamma = 0.2$). Query sizes for the four main datasets are shown below; other parameters are in the Appendix.

\begin{center}
\begin{tabular}{lccc}
\toprule
\textbf{Dataset} & \textbf{NORMAL} & \textbf{OOD} (0.6) & \textbf{Imb} ($\gamma=0.2$) \\
\midrule
CIFAR-10      & 500   & 100 & 100 \\
CIFAR-100     & 1000  & 500 & 500 \\
SST-5         & 10    & 10  & 10  \\
YELP-3000    & 10    & 10  & 10  \\
\bottomrule
\end{tabular}
\label{tab:query_sizes}
\end{center}

The four datasets above are widely used in prior active learning research \cite{tan2023bayesian, parvaneh2022active, ning2022active, bu2018active} and are selected for the main comparison in Table~\ref{table:main_results}. Based on the standard practices adopted in these works, we set our own query batch sizes to ensure fair and consistent evaluation.

For the OOD setting, 60\% of unlabeled data is from unknown classes, simulating a challenging open-set scenario. For the imbalanced setting, $\gamma = 0.2$ means the smallest class contains 20\% as many samples as the largest (see Section~\ref{real_world}). These settings balance realism and difficulty in open-set and imbalanced-set.

All results report mean accuracy at the 10th round, averaged over five trials. Main results on four representative datasets are shown in Table~\ref{table:main_results}, while full results covering all seven datasets and settings are provided in the Appendix. All experiments were run on A100 or RTX 4090 GPUs under consistent conditions.

\subsection{Performance Analysis Across Different Settings}

\paragraph{Results on Image Datasets}
On the CIFAR-10 dataset, uncertainty-based approaches demonstrate superior performance, with CONF and VarRatio attaining the highest accuracy (0.833), closely followed by Margin (0.832).

In contrast, on the more complex CIFAR-100 dataset, which comprises 100 classes, hybrid methods exhibit stronger competitiveness. BADGE achieves the highest accuracy under standard settings (0.578), followed by NoiseStability (0.575). The overall decline in performance relative to CIFAR-10 underscores the increased difficulty of fine-grained multi-class classification, wherein hybrid strategies offer enhanced robustness.

Moreover, the performance disparities among different methods are more pronounced on CIFAR-100, suggesting that the choice of algorithm becomes increasingly consequential with rising task complexity. The observed advantage of hybrid methods may be attributed to their improved capacity to balance exploration and exploitation in high-dimensional, multi-class settings.

\paragraph{Results on Text Datasets}
Text datasets exhibit markedly different performance characteristics compared to image datasets. On SST-5, MeanSTD achieves the best performance under standard settings with an accuracy of 0.352, while CONF and VarRatio tie for second place (0.340). The generally lower absolute accuracy values reflect the inherent complexity of sentiment analysis tasks, where linguistic nuances and contextual dependencies pose greater challenges for active learning algorithms.

The YELP-3000 dataset demonstrates different trends, with CoresetCB leading under standard settings (0.408), followed by CONF and VarRatio (0.407). The superior performance of CoresetCB suggests that methods specifically designed for handling class imbalance may provide advantages even in standard settings for text classification tasks.

\paragraph{OOD Setting Results}
The results under OOD setting diverges notably from standard settings, highlighting the impact of distribution shift on active learning. On CIFAR-10, Random (0.570) and Margin (0.565) outperform more complex uncertainty-based methods, suggesting diversity-based strategies may be more robust OOD.

On CIFAR-100, LfOSA achieves the best result (0.420), affirming the value of aligning algorithms with task-specific challenges like OOD detection. However, LfOSA shows modest performance in other open-set experiments, which may be attributed to significantly smaller query sizes compared to CIFAR-100 settings.

Text datasets exhibit different patterns under OOD conditions. On SST-5, Coreset and CoresetCB achieve the best performance (0.658), indicating that in open-set scenarios, strategies that evaluate data value based on sample diversity also hold significant importance. On YELP-3000, CoreLog leads with the highest performance (0.695). 

\paragraph{Imbalanced Setting Results}
The results under imbalanced setting highlight the importance of methods specifically designed to address class imbalance. On CIFAR-10, TiDAL achieves the best performance (0.505), followed by BADGE (0.503), indicating that hybrid methods combining uncertainty and diversity factors are beneficial for handling imbalanced scenarios.

On CIFAR-100, CoresetCB attains the highest accuracy (0.412), followed by standard Coreset (0.409), indicating that class-balanced variants of core-set methods effectively address imbalance issues. The strong performance of diversity-based sampling methods in this task underscores the importance of ensuring representative sampling of minority classes for imbalanced active learning.

In text imbalanced datasets, results show interesting variations: BALD performs best on SST-5 (0.365), while MeanSTD leads on YELP-3000 (0.460). This indicates that under imbalanced setting, various DAL algorithms exhibit significant performance differences across different datasets. The textual characteristics of the datasets—such as vocabulary diversity, document length, and class distribution—have a substantial impact on the effectiveness of these algorithms.

\subsection{Selection Time Analysis}
\label{sec:time}
To evaluate the computational efficiency of various DAL algorithms during the sample selection phase, we conducted a systematic comparison of the average total selection time per trial for 19 algorithms on the CIFAR-100 dataset, with a query size of 500 samples per round. This experiment was conducted on a single RTX 4090 GPU. The results are presented in Figure \ref{fig:runtime}, and additional results and findings are provided in the Appendix. 

As illustrated in Figure~\ref{fig:runtime}, in experiments conducted on the CIFAR-100 dataset with a query size of 500, the average sample selection time per trial for the majority of DAL algorithms remains below 500 seconds. However, certain complex strategies that incorporate diversity modeling, gradient estimation, or feature fusion exhibit substantially higher computational overhead, notably NoiseStability (10,328.80 seconds), EntropyCB (22,363.02 seconds), and BADGE (62,747.00 seconds).

These findings suggest that, in practical applications, the selection of DAL algorithms should not only consider their ability to reduce annotation costs while preserving model performance, but also account for the computational efficiency of the sample selection phase. This consideration becomes particularly critical in scenarios involving strict time constraints or high system complexity, where the overhead of sample selection may significantly affect overall system efficiency.

\begin{figure}[!t]
    \centering
    \includegraphics[width = \columnwidth]{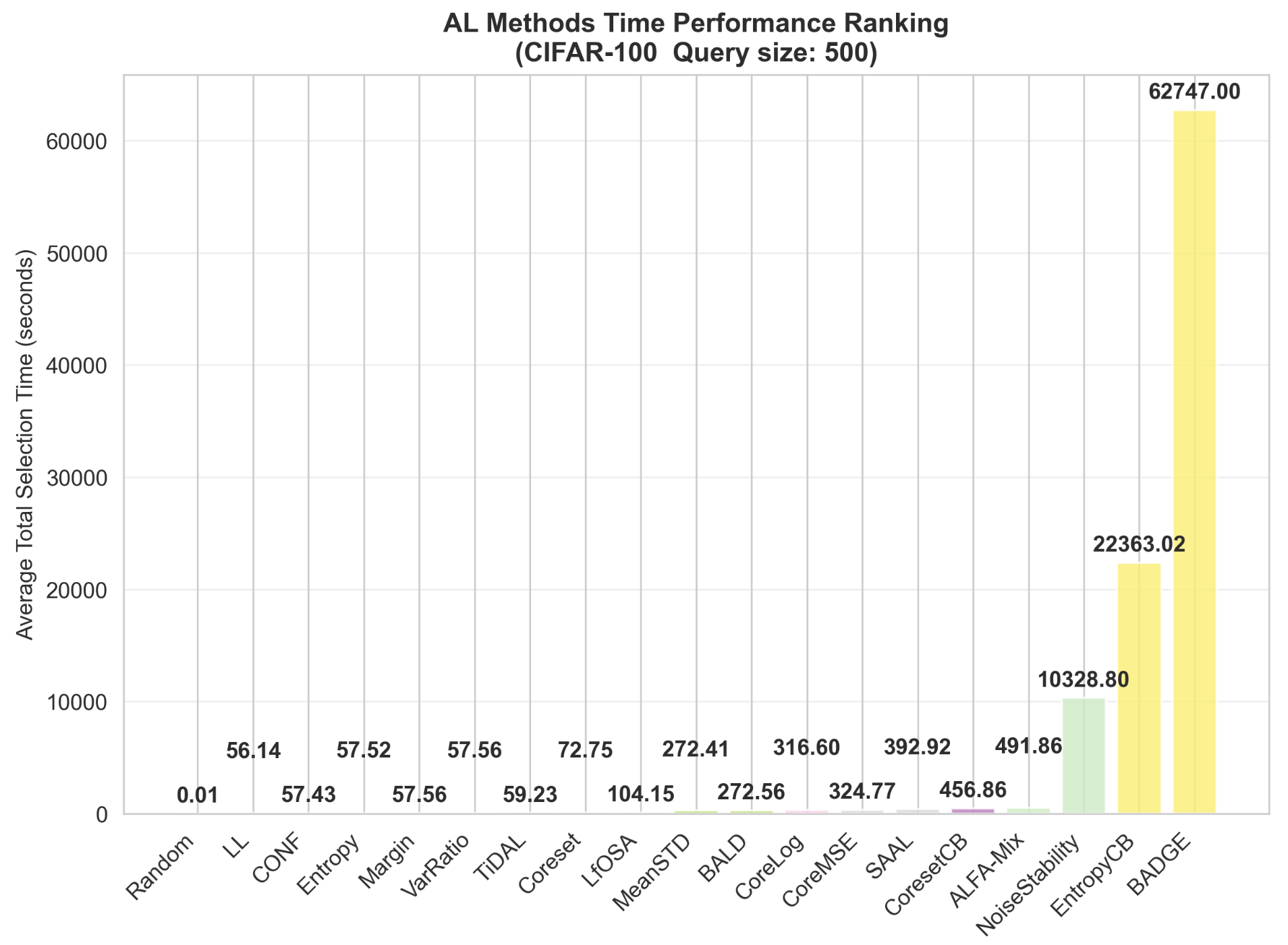}
    \caption{The average runtime per trial for the 19 AL algorithms on the CIFAR-100 dataset (with a query size of 500).}
    \label{fig:runtime}
\end{figure}

\subsection{Statistical Performance Comparison}
To evaluate the statistical performance of various DAL algorithms in Table~\ref{table:main_results}, we adopted the method from \cite{tan2023bayesian}, using a two-sided paired t-test based on accuracy at ten evenly spaced sample points from the learning curve. These points correspond to the 1st through 10th cycles with a step size of 1.

For each experiment setting (same dataset, seed, and hyperparameters), we performed pairwise comparisons between methods $i$ and $j$. The t-score was computed as:
$$
t = \frac{\bar{x}}{s / \sqrt{10}}, \quad \bar{x} = \frac{1}{10} \sum_{k=1}^{10} l_k^{ijd}, \quad s = \sqrt{ \frac{1}{9} \sum_{k=1}^{10} (l_k^{ijd} - \bar{x})^2 }
$$

\begin{figure}[!t]
    \centering
    \includegraphics[width = \columnwidth]{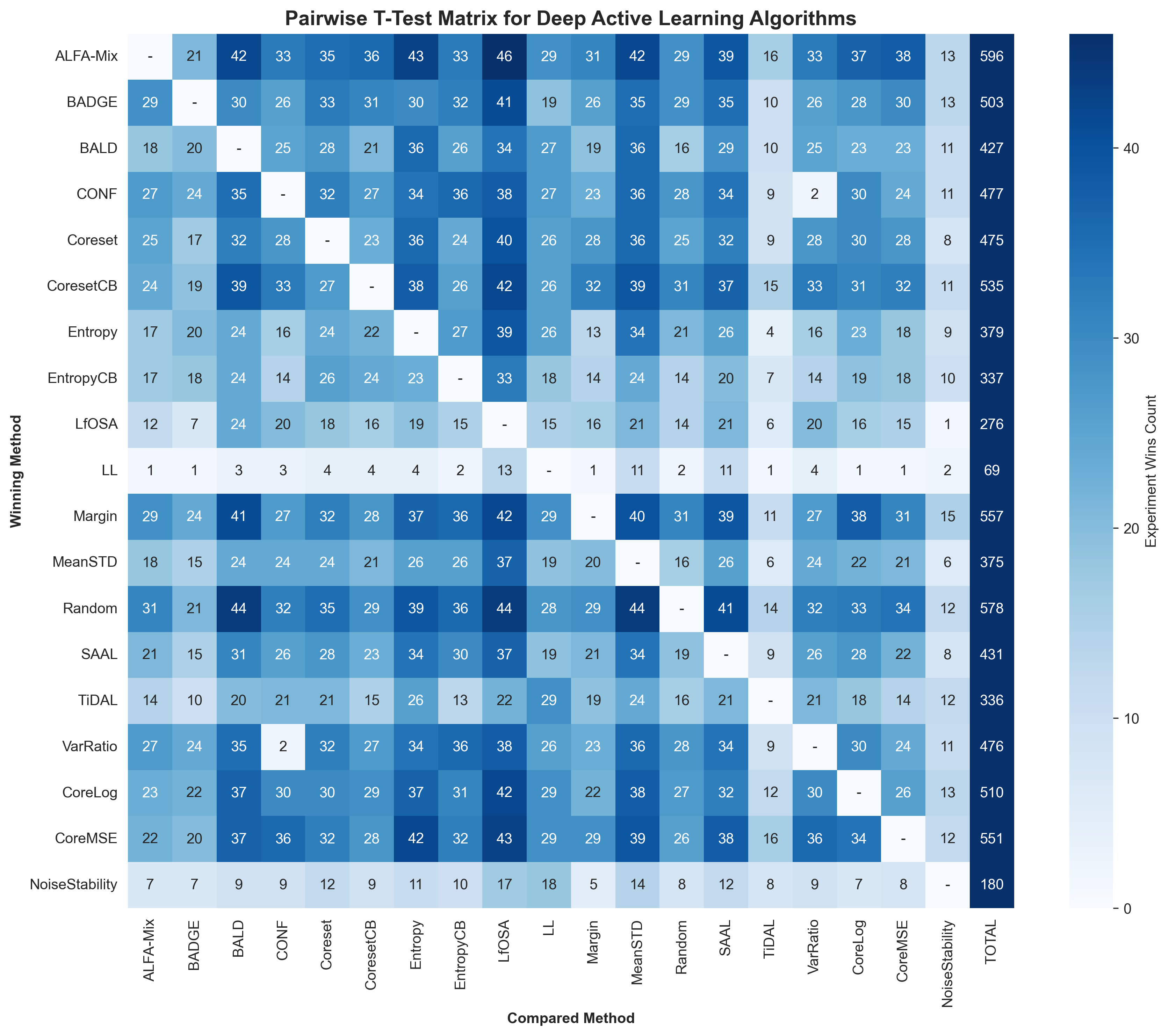}
    \caption{Accuracy-based pairwise comparison for experiments in Table~\ref{table:main_results}.}
    \label{fig:matrix}
\end{figure}

When $p < 0.05$, method $i$ is considered to significantly outperform method $j$ within that setting. Based on this criterion, we constructed a pairwise comparison matrix $C$ (Figure~\ref{fig:matrix}), where $C_{ij}$ represents the number of experiment pairs in which method $i$ outperforms method $j$ across all settings. The “Total” column represents the total number of experiment-level wins each method achieves across all pairwise comparisons.

Figure~\ref{fig:matrix} presents the key findings: ALFA-Mix leading with 596 wins, making it the most consistently effective method. Random sampling follows closely with 578 wins, underscoring the strength of simple baselines. Among DAL algorithms, CoreMSE (551 wins) performs well, while CONF (477 wins) shows moderate success. Entropy (379 wins) perform relatively poorly, and specialized techniques like LL (69 wins) show limited effectiveness. We present additional comparison matrices under various settings in the Appendix.

\section{Conclusion}
We present ALScope, a novel DAL platform that enables consistent evaluation across widely-used image and text classification datasets, while streamlining the development and assessment of DAL algorithms. Leveraging ALScope, we conduct a comprehensive comparative study of up to 19 DAL algorithms across 7 publicly available datasets, yielding the following key observations: 
(1) DAL algorithms exhibit significant performance variation across different domains and task configurations. 
(2) Existing DAL methods show clear limitations in non-standard scenarios such as class imbalance and open-set conditions, highlighting the need for further exploration. 
(3) Although some algorithms perform well, they incur considerable selection-time overhead, requiring practitioners to balance accuracy and efficiency. 
Furthermore, we will continue to expand ALScope by incorporating more algorithms and extending its support to broader application domains.

{\small
\bibliography{aaai2026}
}

\clearpage
\appendix
\section{Hardware and Environment}
\label{sec:env}
Most experiments were conducted on A100 GPUs. The following experiments, however, were performed on a single RTX 4090:
\begin{itemize}
  \item \textbf{CIFAR-10 (Openset)}: OOD Ratio = 0.2,\ 0.4
  \item \textbf{CIFAR-100 (Openset)}: OOD Ratio = 0.6
  \item \textbf{SST-5 (Openset)}: OOD Ratio = 0.2,\ 0.4,\ 0.6
  \item \textbf{Yelp (Imbalance)}: $\gamma$ = 0.2,\ 0.4,\ 0.6
  \item \textbf{Yelp (Openset)}: OOD Ratio = 0.2,\ 0.4,\ 0.6
  \item \textbf{Selection Runtime Statistics}
\end{itemize}

ALScope was implemented on Ubuntu 22.04 using NVIDIA CUDA 12.2 and PyTorch 2.2.2. The complete Python dependencies are provided in requirements.txt and EnvironmentSetup.md in the Code Appendix. 

\section{Hyperparameter Configuration}
\label{sec:hyperparameter_config}

This section provides a comprehensive overview of the hyperparameters employed in our active learning framework. The parameters are organized into distinct categories based on their functional roles within the experimental pipeline.

\subsection{Key Parameters in Experiments}

Here are the key parameters used in our experiments:

\textbf{Backbone Models:} We employ DistilBert for text classification tasks and ResNet18 for image classification. 

\textbf{Random Seed:} Seeds ranging from 0-4 are used to ensure reproducibility across multiple experimental trials, allowing for statistical significance testing and variance analysis.

\textbf{Optimizers:} AdamW is used for text models, combining the benefits of Adam optimization with decoupled weight decay regularization. SGD (Stochastic Gradient Descent) with momentum is employed for image models, which often performs well with convolutional architectures.

\textbf{Learning Rate:} The learning rate controls the step size during gradient descent optimization. Text models use $1 \times 10^{-4}$, while image models use 0.1.

\textbf{Weight Decay:} Set to $5 \times 10^{-4}$ for both modalities, this L2 regularization parameter prevents overfitting by penalizing large weights in the model.

\textbf{Training Epochs:} Text models are trained for 30 epochs per cycle, while image models require 200 epochs.

\textbf{Learning Rate Scheduler:} StepLR scheduler reduces the learning rate by a factor of gamma (0.5) every step\_size epochs (50).

\textbf{Active Learning Configuration:} In each set of experiments, for each type of DAL algorithm, we conduct 5 trials, and each trial consists of 10 active learning cycles. The initial training set size is kept equal to the query size, and Monte Carlo dropout with n-drop=5 is used when applicable for uncertainty estimation.

\subsection{Experimental Setup Parameters}
\label{subsec:experimental_setup}

The foundational experimental configuration includes dataset specification (\textbf{--dataset}), model architecture selection (\textbf{--model}), and computational resource allocation (\textbf{--gpu}, \textbf{--data-parallel}). The experimental robustness is ensured through multiple trial runs (\textbf{--trial}, default: 5) with controlled randomization via seeded initialization (\textbf{--seed}, default: 0). Dataset-specific parameters include the number of classes (\textbf{--n-class}), image resolution (\textbf{--resolution}), and data loading specifications (\textbf{--workers}, \textbf{--data\_path}).

\subsection{Active Learning Framework Parameters}
\label{subsec:al_framework}

The core active learning process is governed by several key parameters. The initial labeled set size is controlled by \textbf{--n-initial} (default: 100), while the iterative learning process is defined by the number of acquisition cycles (\textbf{--cycle}, default: 10) and query batch size (\textbf{--n-query}). The acquisition strategy is specified through \textbf{--method}, supporting various approaches including uncertainty-based sampling, coreset selection, and advanced methods such as BADGE, CCAL, and MQNet. Method-specific configurations include classic uncertainty-based measures (\textbf{--uncertainty}) and submodular optimization parameters (\textbf{--submodular}, \textbf{--submodular\_greedy}).

\subsection{Data Configuration and Preprocessing}
\label{subsec:data_config}

Data handling encompasses balanced subset selection through \textbf{--samples-per-class} and \textbf{--subset-preset} options, enabling controlled dataset scaling. The framework supports challenging scenarios including open-set recognition (\textbf{--openset}), imbalanced classification (\textbf{--imb-factor}), and OOD detection (\textbf{--ood-rate}). Text dataset compatibility is provided through \textbf{--textset}.

\subsection{Training and Optimization Parameters}
\label{subsec:training_optimization}

The training regimen is configured through epoch specifications (\textbf{--epochs}, default: 200) with method-specific variants for specialized components (\textbf{--epochs-ccal}, \textbf{--epochs-csi}, \textbf{--epochs-mqnet}). Optimization parameters include learning rate schedules (\textbf{--lr}, \textbf{--scheduler}), momentum (\textbf{--momentum}), and weight decay (\textbf{--weight\_decay}). Batch size configurations are provided for different training phases (\textbf{--batch-size}, \textbf{--ccal-batch-size}, \textbf{--csi-batch-size}) along with testing parameters (\textbf{--test-batch-size}, \textbf{--test\_interval}).

\subsection{Advanced Method-Specific Parameters}
\label{subsec:advanced_methods}

\subsubsection{Sharpness-Aware Miniaturization (SAM)}
\label{subsubsec:sam}

SAM-based acquisition modes are controlled through \textbf{--acqMode} (Max\_Diversity, Diff\_Diversity) and \textbf{--labelMode} (True, Pseudo, InversePseudo), with sharpness computation regulated by the perturbation radius \textbf{--rho} (default: 0.05).

\subsubsection{ALFA-Mix Configuration}
\label{subsubsec:alphamix}

The ALFA-Mix method employs several specialized parameters including \textbf{--alpha\_cap} (default: 0.03125), optimization flags (\textbf{--alpha\_opt}, \textbf{--alpha\_closed\_form\_approx}), and learning parameters (\textbf{--alpha\_learning\_rate}, \textbf{--alpha\_learning\_iters}) for adaptive mixing coefficient determination.

\subsubsection{Contrastive and Self-Supervised Learning (CCAL)}
\label{subsubsec:ccal}

CCAL implementation includes data augmentation parameters (\textbf{--resize\_factor}, \textbf{--resize\_fix}), contrastive loss weighting (\textbf{--sim\_lambda}), and transformation specifications (\textbf{--shift\_trans\_type}). OOD scoring is configured through \textbf{--ood\_samples} and temperature parameters (\textbf{--k}, \textbf{--t}).

\subsubsection{Noise Stability Analysis}
\label{subsubsec:noise_stability}

Noise stability evaluation employs Monte Carlo sampling (\textbf{--noise\_sampling}, default: 50) with controlled perturbation scaling (\textbf{--noise\_scale}, default: 0.001).

\subsection{Computational and Storage Parameters}
\label{subsec:computational_storage}

Computational efficiency is managed through chunking strategies (\textbf{--chunk\_size}), pool subset limitations (\textbf{--pool\_subset}, \textbf{--pool\_batch\_size}), and parallel processing configurations. Model persistence is controlled via checkpoint parameters (\textbf{--save\_path}, \textbf{--resume}) with optional SSL model saving (\textbf{--ssl\_save}).

\subsection{Evaluation and Reporting Parameters}
\label{subsec:evaluation_reporting}

The evaluation framework includes test set configuration (\textbf{--test\_fraction}), performance monitoring intervals (\textbf{--print\_freq}), and detailed logging options (\textbf{--verbose-subset}). Method-specific evaluation parameters ensure comprehensive assessment across different active learning strategies.

The extensive parameterization enables systematic exploration of the active learning design space while maintaining experimental reproducibility and methodological rigor. Default values are established based on empirical best practices and preliminary experiments, with parameter sensitivity analysis conducted where applicable.

\section{Experiment Scripts and Logs}
\label{sec:scripts_logs}
\subsection{Command Lines and Scripts}
When conducting DAL experiments with ALScope, users can conveniently specify the required experimental parameters via the command line and initiate the experiment accordingly. Clearly, multiple command-line instructions can also be consolidated into a script file to facilitate batch execution of experiments. Below are representative command-line examples along with their corresponding explanations.

\paragraph{Standard Setting} This is a standard DAL setup:

\begin{itemize}
\item python main.py --method coremse --uncertainty Entropy --dataset CIFAR100 --trial 5 --cycle 10 --n-initial 500 --n-query 500 --n-class 100 --epochs 200 --n-drop 5
\end{itemize}

Key parameters include:
\begin{itemize}
    \item \textbf{--method}: Specifies the DAL algorithm.
    \item \textbf{--uncertainty}: Specifies one of the classical uncertainty-based DAL algorithms.
    \item \textbf{--dataset}: Specifies the dataset.
    \item \textbf{--trial}: Experiment trial/run number.
    \item \textbf{--cycle}: Total number of active learning cycles to perform in each trial.
    \item \textbf{--n-initial}: Initial number of labeled samples to start with.
    \item \textbf{--n-query}: Number of samples to query and label in each active learning cycle.
    \item \textbf{--n-class}: Number of classes in the dataset.
    \item \textbf{--epochs}: Number of training epochs per active learning cycle.
    \item \textbf{--n-drop}: Number of Monte Carlo Dropout iterations for uncertainty estimation (if applicable).
\end{itemize}

\paragraph{Open-set Setting} This setting is for Open-set Active Learning, where the unlabeled pool includes unknown (OOD) samples:

\begin{itemize}
\item python main.py --method coremse --uncertainty Entropy --dataset CIFAR10 --trial 5 --cycle 10 --n-initial 100 --n-query 100 --n-class 10 --epochs 200 --n-drop 5 --openset --ood-rate 0.2
\end{itemize}

In addition to the standard parameters, the following are used:
\begin{itemize}
    \item \textbf{--openset}: Enables open-set recognition mode.
    \item \textbf{--ood-rate}: Specifies that percentage of the unlabeled pool consists of OOD samples to simulate real-world noise or unknown classes.
\end{itemize}

\paragraph{Imbalanced-set Setting} This is for Imbalanced Active Learning, designed to simulate scenarios with severe class imbalance:

\begin{itemize}
\item python main.py --method coremse --uncertainty Entropy --dataset CIFAR10 --trial 5 --cycle 10 --n-initial 100 --n-query 100 --n-class 10 --epochs 200 --n-drop 5 --imb-factor 0.2
\end{itemize}

Compared to the standard setting, it adds:
\begin{itemize}
    \item \textbf{--imb-factor}: Enables Imbalanced-set mode and specifies the class imbalance factor (i.e., $\gamma$); a lower value (e.g., 0.2) indicates a highly long-tailed distribution.
\end{itemize}

\subsection{Experimental Log}
\begin{figure*}[!ht]
    \centering
    \includegraphics[width = \textwidth]{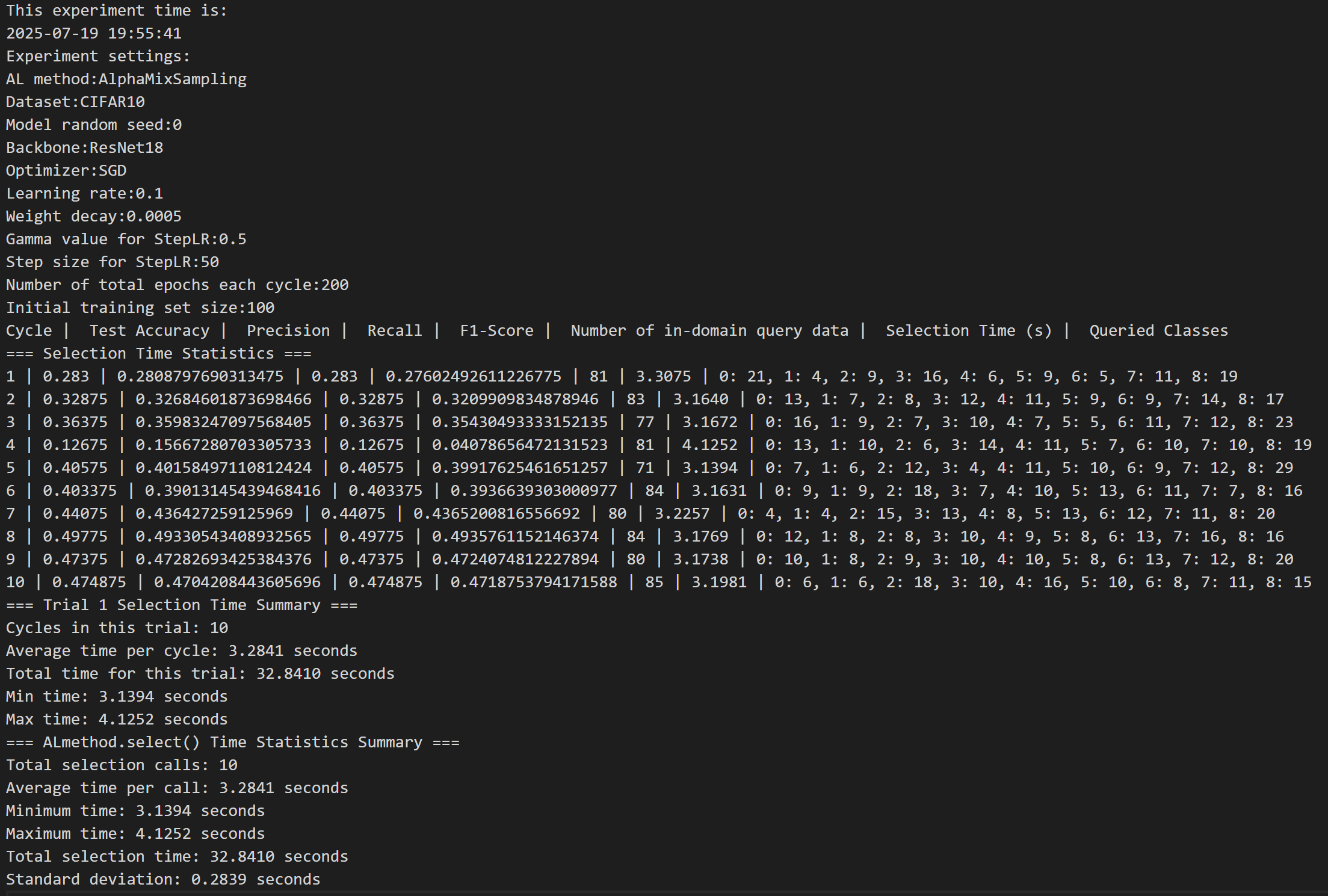}
    \caption{A screenshot of an experiment log.}
    \label{fig:log}
\end{figure*}

Our experimental log records the configuration and per-cycle results of a single experiment using a specific DAL algorithm on a given dataset. The log includes basic configuration details such as the model’s random seed, backbone network, optimizer settings, learning rate scheduling strategy, number of training epochs per cycle, and the initial training set size. For each active learning cycle, the log documents the test accuracy, precision, recall, F1-score, number of queried samples, selection time, and the classes of the selected samples. It also provides summary statistics on the selection time. Figure~\ref{fig:log} is an example.

\section{Details of Implemented DAL Algorithms in ALScope}
\label{all_algorithms}

\paragraph{CONF} The Least Confidence method, proposed by Lewis and Gale \cite{lewis1995sequential}, selects samples with the lowest predicted confidence, i.e., those for which the model is most uncertain.

\paragraph{Entropy} Based on Shannon’s information theory \cite{shannon2001mathematical}, this method uses the entropy of the predictive distribution as an uncertainty metric—higher entropy indicates higher uncertainty.

\paragraph{Margin} Scheffer et al. \cite{scheffer2001active} introduced margin sampling, which measures uncertainty by the difference between the top two predicted class probabilities; a smaller margin implies greater uncertainty.

\paragraph{MeanSTD} Kampffmeyer et al. \cite{kampffmeyer2016semantic} proposed this method, which uses the standard deviation across multiple predictions to assess consistency; larger deviations indicate higher uncertainty.

\paragraph{BALD} Gal et al. \cite{gal2017deep} proposed Bayesian Active Learning by Disagreement (BALD), which employs Bayesian neural networks and measures mutual information between predictions and model parameters to quantify epistemic uncertainty.

\paragraph{VarRatio} The variation ratio, proposed by Freeman \cite{freeman1965elementary}, calculates uncertainty based on the frequency of the most common prediction among model trials—a lower ratio indicates greater uncertainty.

\paragraph{LL} Yoo and Kweon \cite{yoo2019learning} introduced the Learning Loss method, which trains an auxiliary network to predict the training loss of unlabeled samples, thus estimating their informativeness.

\paragraph{Coreset} Sener and Savarese \cite{sener2017active} proposed the Core-set method, modeling sample selection as a maximum coverage problem to choose a representative and diverse subset of the data.

\paragraph{ALFA-Mix} Parvaneh et al. \cite{parvaneh2022active} proposed this method, which combines mixup data augmentation with uncertainty and diversity metrics to select highly informative synthetic samples.

\paragraph{BADGE} Ash et al. \cite{ash2019deep} proposed BADGE, a hybrid method that combines uncertainty and gradient-based diversity by clustering gradient embeddings using k-means++.

\paragraph{CoreLog} Tan et al. \cite{tan2023bayesian} introduced CoreLog, a Bayesian active learning method that integrates core-set selection with logit-based uncertainty, particularly effective in imbalanced datasets.

\paragraph{CoreMSE} Also proposed by Tan et al. \cite{tan2023bayesian}, CoreMSE uses mean squared error to evaluate prediction uncertainty and constructs an informative core-set accordingly.

\paragraph{SAAL} Kim et al. \cite{kim2023saal} proposed SAAL (Sharpness-Aware Active Learning), which selects informative samples based on sharpness-aware loss perturbations. By leveraging pseudo-labels and estimating perturbed losses, SAAL identifies samples with high uncertainty and potential impact. This approach improves label efficiency across various vision tasks.

\paragraph{NoiseStability} Li et al. \cite{li2024deep} introduced NoiseStability, which evaluates the informativeness of a sample by measuring prediction robustness under perturbations—less stable samples are considered more informative.

\paragraph{TiDAL} Kye et al. \cite{kye2023TiDAL} proposed TiDAL (Training Dynamics for Active Learning), which identifies uncertain samples by predicting their training dynamics using a separate module. It selects samples with unstable prediction trajectories, capturing long-term uncertainty without requiring adversarial perturbations or multiple forward passes.

\paragraph{CoresetCB} Bengar et al. \cite{bengar2022class} proposed CoresetCB, an extension of core-set selection that accounts for class imbalance during representative subset construction.

\paragraph{EntropyCB} Also introduced by Bengar et al. \cite{bengar2022class}, EntropyCB adapts entropy-based sampling to handle class imbalance by incorporating class frequency into the selection process.

\paragraph{CCAL} Du et al. \cite{du2021contrastive} proposed Contrastive Active Learning (CCAL), which enhances feature discrimination via contrastive learning to improve performance in open-set scenarios.

\paragraph{LfOSA} Ning et al. \cite{ning2022active} proposed Learning from Open-set Active data (LfOSA), which targets the identification of unknown-class samples and leverages both known and unknown data for selection.

\paragraph{MQ-Net} Park et al. \cite{park2022meta} proposed MQ-Net, a meta-learning-based framework designed to handle open-set active learning and improve unknown class recognition.

\paragraph{Random} Random sampling is a baseline method that selects samples without considering uncertainty or diversity.

\section{Additional Selection Runtime Analysis}
\label{sec:time_continue}

\begin{figure}[!ht]
    \centering
    \includegraphics[width = \columnwidth]{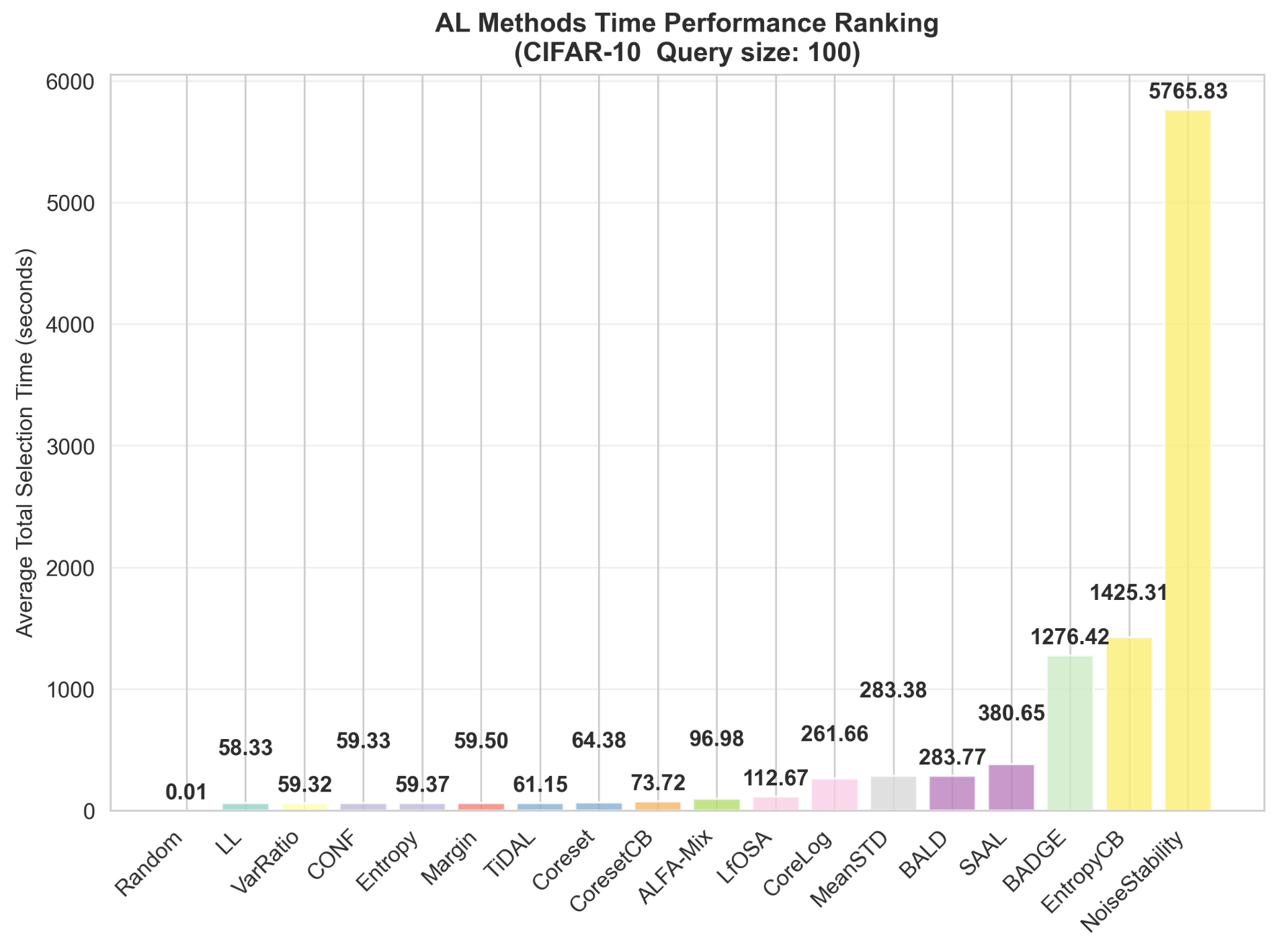}
    \caption{The average runtime per trial for the 19 AL algorithms on the CIFAR-10 dataset (with a query size of 100).}
    \label{fig:cifar10_time_1}
\end{figure}

\begin{figure}[!ht]
    \centering
    \includegraphics[width = \columnwidth]{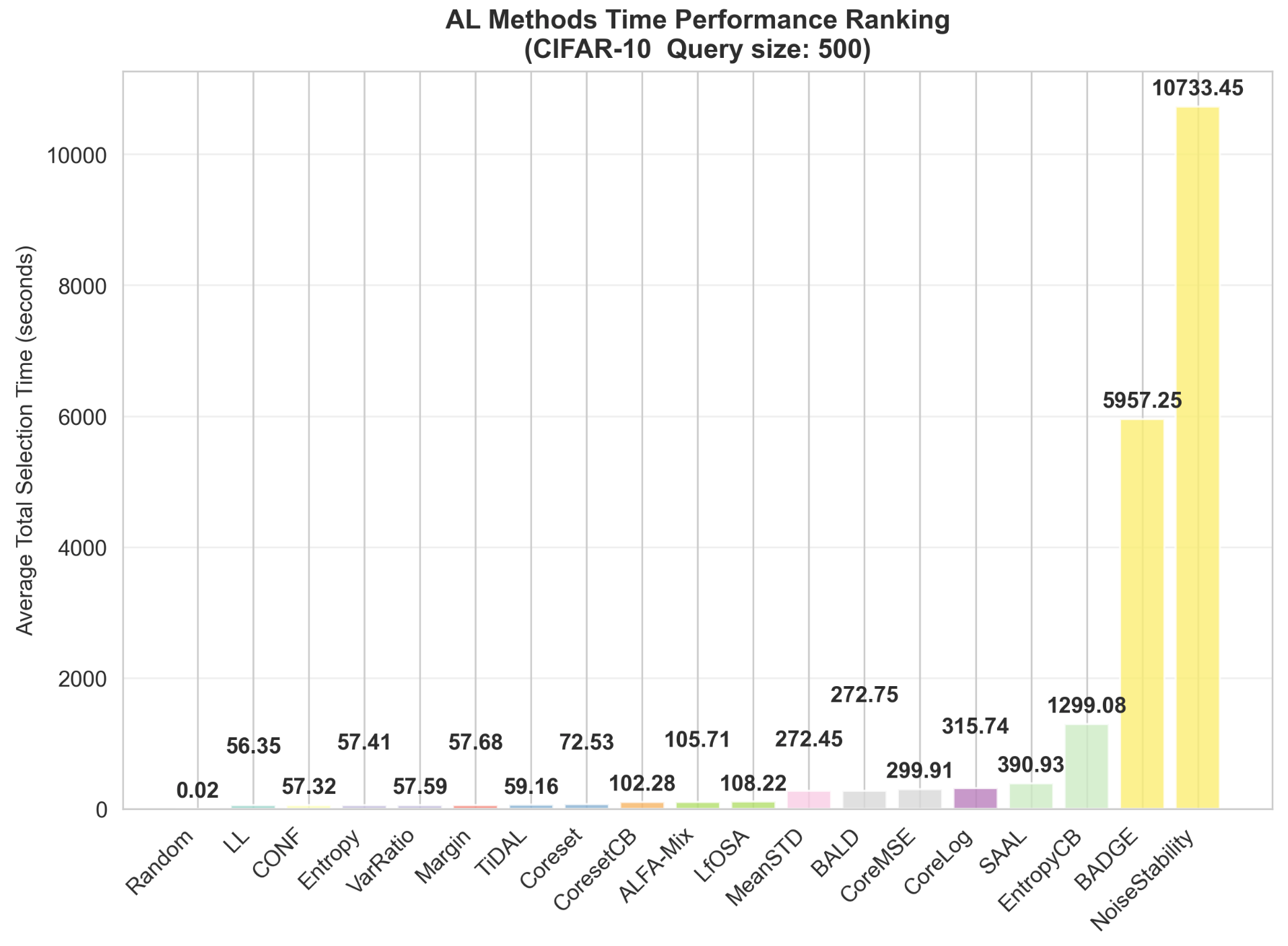}
    \caption{The average runtime per trial for the 19 AL algorithms on the CIFAR-10 dataset (with a query size of 500).}
    \label{fig:cifar10_time_2}
\end{figure}

In addition to the selection runtime experiments conducted on the CIFAR-100 dataset, we also performed analogous experiments on CIFAR-10, considering two different query sizes: 100 (Figure~\ref{fig:cifar10_time_1}) and 500 (Figure~\ref{fig:cifar10_time_2}). Across all scenarios—including those previously reported on CIFAR-100—the three most time-consuming algorithms consistently were EntropyCB, BADGE, and NoiseStability. Furthermore, for a fixed dataset, increasing the query size results in longer selection runtimes. Similarly, switching from the simpler CIFAR-10 dataset to the more complex CIFAR-100 dataset (with a larger number of classes), while keeping the query size fixed, also leads to increased selection runtime. However, the degree of sensitivity to these two factors—query size and dataset complexity—varies across different selection algorithms.

For instance, on the CIFAR-10 dataset with a query size of 100, ALFA-Mix exhibits an average selection runtime of 96.98 seconds per trial (Figure~\ref{fig:cifar10_time_1}). When the query size increases to 500, the runtime increases slightly to 105.71 seconds Figure~\ref{fig:cifar10_time_2}), indicating limited sensitivity to query size. However, when the dataset is changed to CIFAR-100 with the same query size of 500, the selection runtime increases substantially to 491.86 seconds (Figure 2 in the main text), suggesting that ALFA-Mix is more sensitive to dataset complexity than to query size.

In contrast, BADGE demonstrates a different trend. On CIFAR-10 with a query size of 100, its average selection runtime per trial is 1276.42 seconds (Figure~\ref{fig:cifar10_time_1}), which increases markedly to 5957.25 seconds when the query size is raised to 500 (Figure~\ref{fig:cifar10_time_2}), showing an approximately linear relationship between runtime and query size. Furthermore, on the more complex CIFAR-100 dataset with a query size of 500, the selection runtime reaches 62747.00 seconds (Figure 2 in the main text), indicating that BADGE is highly sensitive to both query size and dataset complexity.

\section{Additional Statistical Performance Comparison}

\begin{figure}[!ht]
    \centering
    \includegraphics[width = \columnwidth]{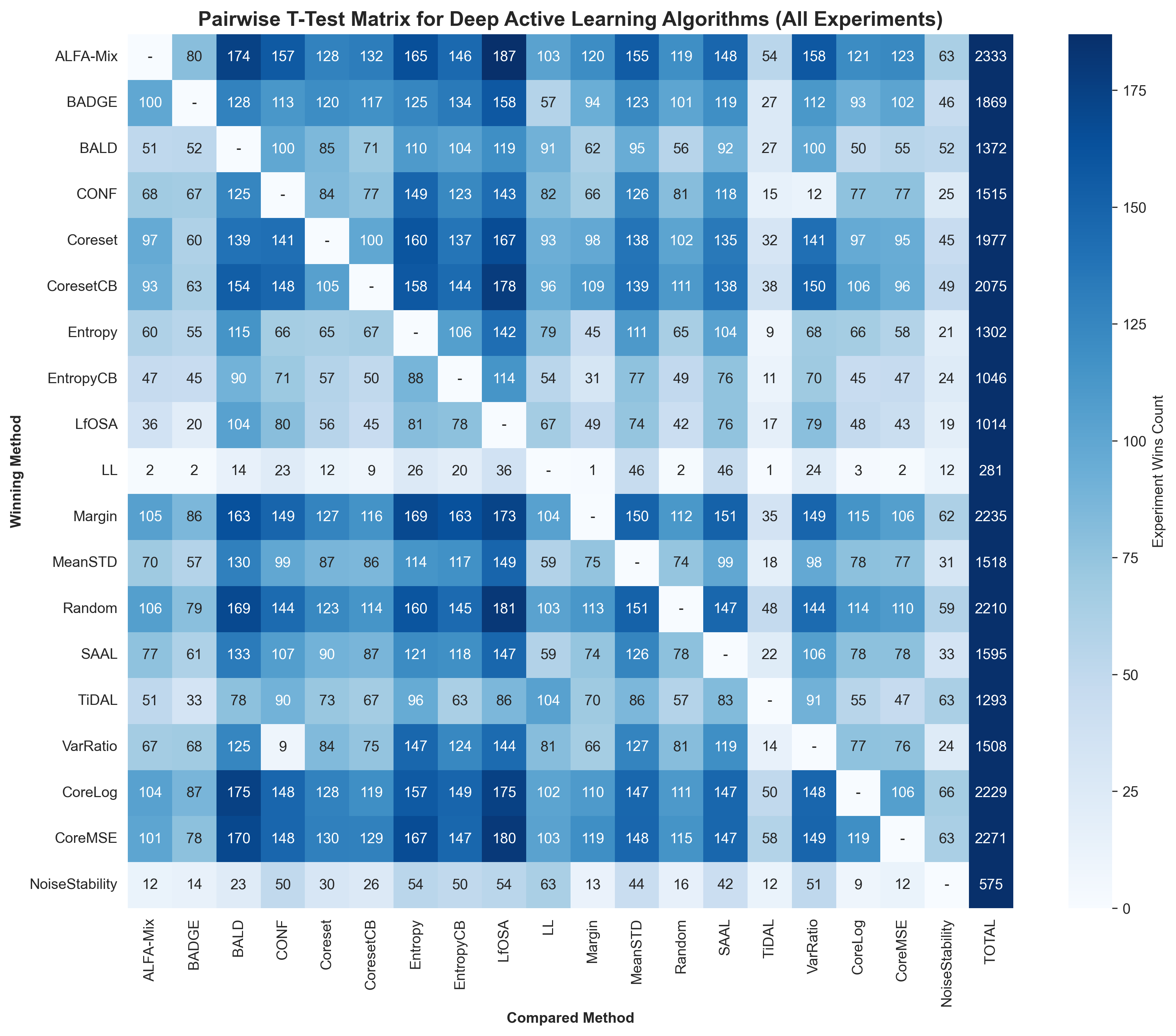}
    \caption{Pairwise T-Test Matrix for DAL algorithms on all experiments.}
    \label{fig:test_all}
\end{figure}

\begin{figure}[!ht]
    \centering
    \includegraphics[width = \columnwidth]{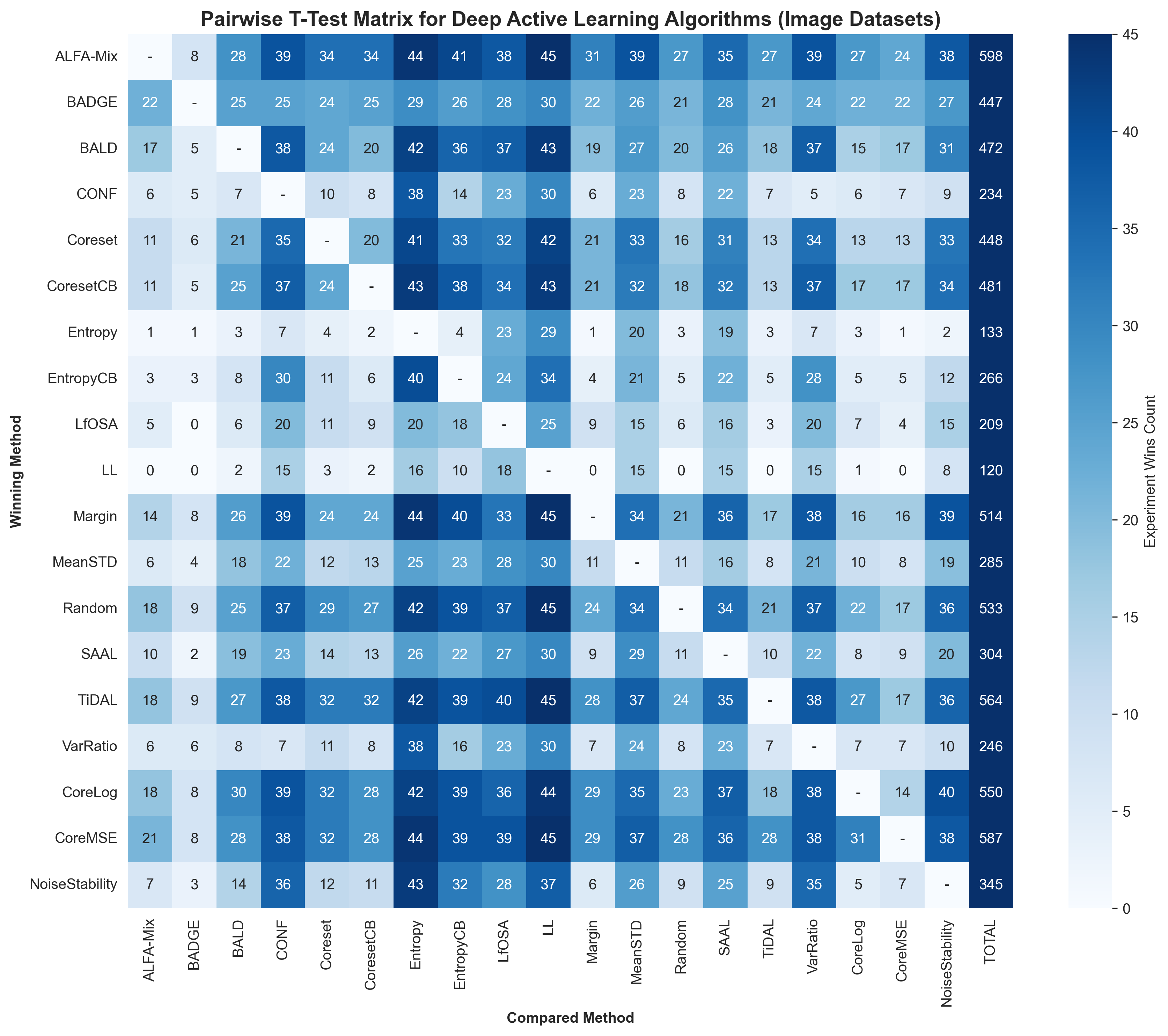}
    \caption{Pairwise T-Test Matrix for DAL algorithms on image datasets.}
    \label{fig:test_image}
\end{figure}

\begin{figure}[!ht]
    \centering
    \includegraphics[width = \columnwidth]{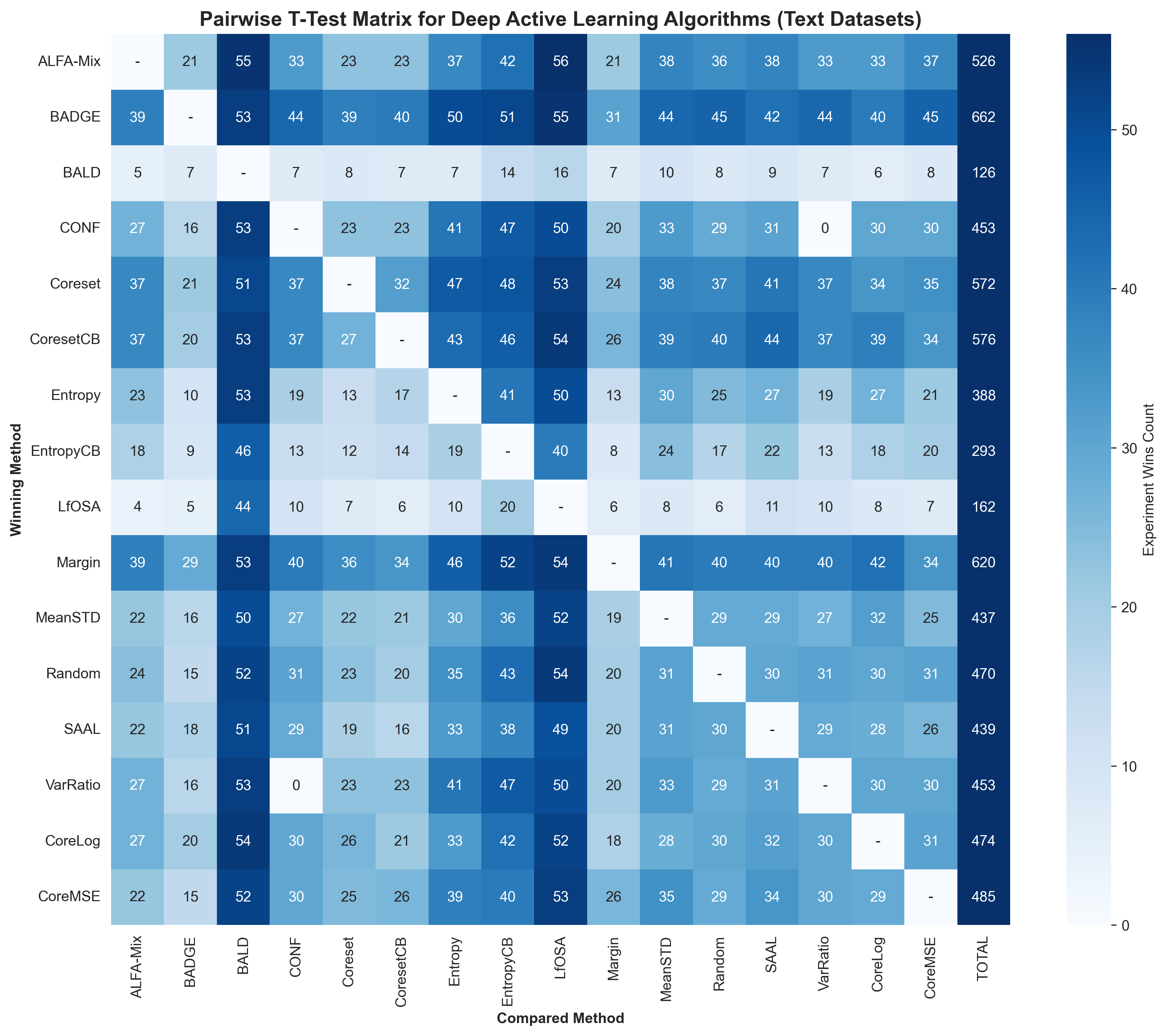}
    \caption{Pairwise T-Test Matrix for DAL algorithms on text datasets.}
    \label{fig:test_text}
\end{figure}

\begin{figure}[!ht]
    \centering
    \includegraphics[width = \columnwidth]{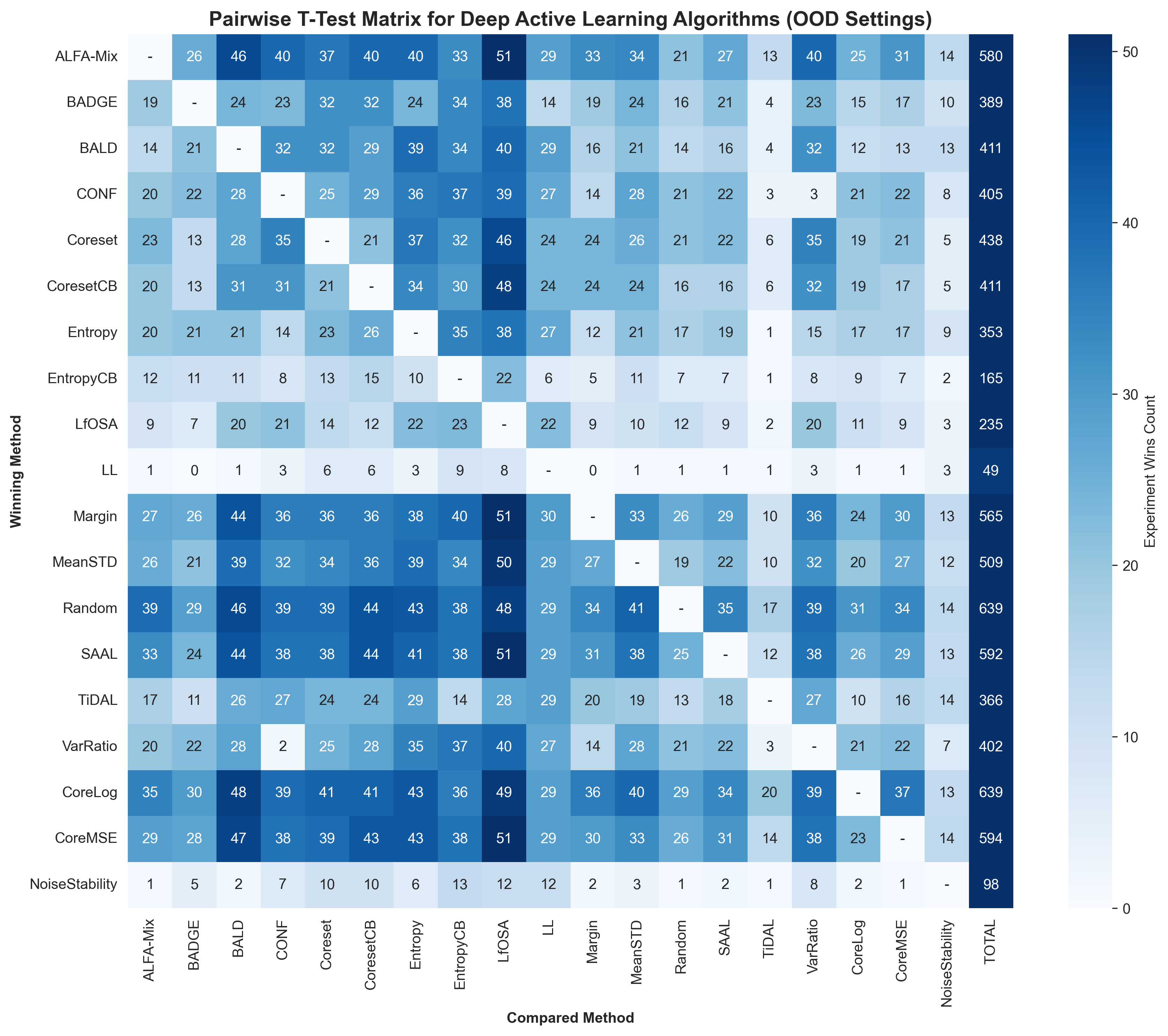}
    \caption{Pairwise T-Test Matrix for DAL algorithms on OOD setting.}
    \label{fig:test_ood}
\end{figure}

\begin{figure}[!ht]
    \centering
    \includegraphics[width = \columnwidth]{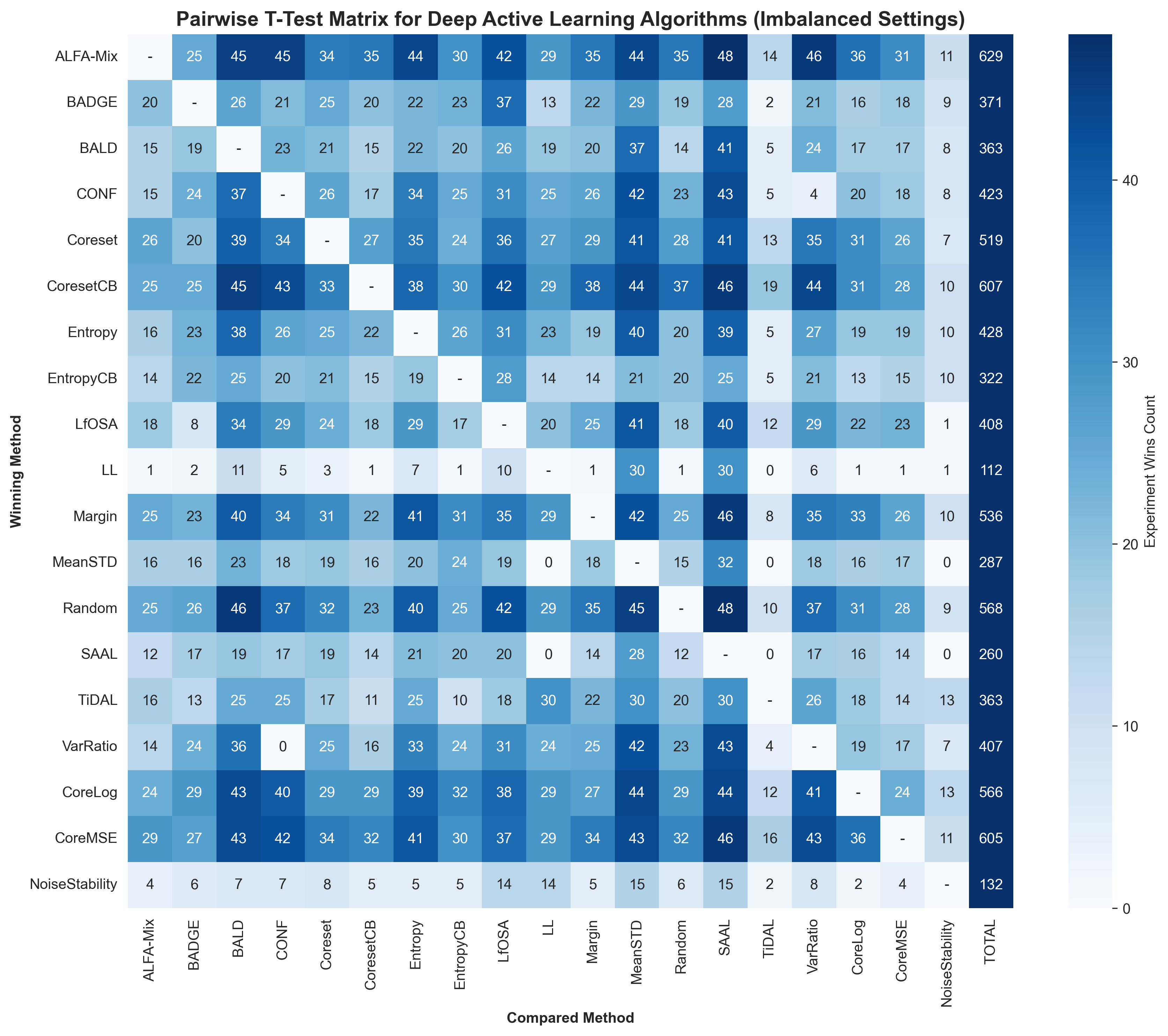}
    \caption{Pairwise T-Test Matrix for DAL algorithms on imbalanced setting.}
    \label{fig:test_imb}
\end{figure}

To provide a comprehensive evaluation of DAL algorithms under diverse experimental conditions, we extend the statistical analysis presented in the main paper to domain-specific and challenging scenarios. Following the methodology detailed in Section 6.3 in the main text, we conduct pairwise t-tests on classification accuracy measured at ten evenly spaced sampling points along the learning curves for each experimental setting.

Figures \ref{fig:test_all} through \ref{fig:test_imb} report the pairwise comparison matrices across five distinct experimental settings: the aggregated results across all experiments, image datasets, text datasets, OOD scenarios, and class-imbalanced datasets. Each matrix $C$ quantifies the number of experiment pairs in which algorithm $i$ significantly outperforms algorithm $j$ (p < 0.05), with the “Total” column indicating the overall number of significant wins for each method.

\paragraph{Overall Performance}
As shown in Figure \ref{fig:test_all}, ALFA-Mix achieves the highest number of significant wins, followed by the CoreMSE and CoreLog variants. This reinforces the efficacy of hybrid strategies that integrate multiple uncertainty estimation techniques.

\paragraph{Standard Settings}

Text Datasets (Figure \ref{fig:test_text}):
Badge sampling ranks highest with 622 wins, followed closely by Margin sampling (620 wins) and CoreSetCB (576 wins). These results suggest that uncertainty-based approaches are particularly well-suited for natural language processing tasks.

Image Datasets (Figure \ref{fig:test_image}):
ALFA-Mix maintains its leading position with 598 wins, followed by CoreMSE (587 wins) and CoreLog (550 wins). The consistent performance of ALFA-Mix across visual domains highlights its robustness in uncertainty quantification.

\paragraph{Real-world Scenarios}
OOD Setting (Figure \ref{fig:test_ood}):
CoreLog achieves the best performance (639 wins), with CoreMSE (594 wins) and SAAL (592 wins) close behind. These results indicate that core-set-based methods are particularly effective in handling distributional shifts and associated uncertainty.

Imbalanced setting (Figure \ref{fig:test_imb}):
ALFA-Mix again achieves the highest number of wins (629), followed by CoreSetCB (607 wins) and CoreMSE (605 wins). This emphasizes the importance of hybrid sampling strategies when dealing with imbalanced label distributions.

\paragraph{Implications}
These results reveals several important trends: (1) No single algorithm consistently dominates across all scenarios, highlighting the necessity of context-aware algorithm selection; (2) Hybrid approaches such as ALFA-Mix exhibit robust and consistent performance across settings; (3) CoreMSE and CoreLog perform strongly under various challenging conditions; (4) Simple baselines such as Margin remain competitive, particularly in text classification tasks.

\section{Full Experimental Results}
\label{sec:all_results}

\begin{table*}[!ht]
    \centering
    \begin{tabular}{lccc|ccc}
    \toprule
    \bfseries Dataset & \multicolumn{3}{c|}{CIFAR-10} & \multicolumn{3}{c}{CIFAR-100} \\
    \midrule
    \bfseries Query Size & 100 & 200 & 500 & 200 & 500 & 1000 \\
     & Acc/AUBC & Acc/AUBC & Acc/AUBC & Acc/AUBC & Acc/AUBC & Acc/AUBC \\
    \midrule
    ALFA-Mix & 0.473 / 0.353 & 0.623 / 0.468 & 0.810 / 0.650 & 0.204 / 0.135 & 0.381 / 0.245 & 0.561 / 0.384 \\
    BADGE & 0.459 / 0.360 & 0.608 / 0.466 & 0.827 / 0.657 & 0.200 / 0.135 & 0.390 / 0.248 & 0.578 / 0.392 \\
    BALD & 0.462 / 0.346 & 0.620 / 0.457 & 0.816 / 0.650 & 0.187 / 0.125 & 0.373 / 0.236 & 0.567 / 0.384 \\
    CONF & 0.422 / 0.327 & 0.610 / 0.452 & 0.833 / 0.664 & 0.168 / 0.110 & 0.369 / 0.227 & 0.566 / 0.370 \\
    Coreset & 0.451 / 0.337 & 0.598 / 0.448 & 0.823 / 0.644 & 0.189 / 0.127 & 0.385 / 0.247 & 0.572 / 0.392 \\
    CoresetCB & 0.432 / 0.341 & 0.599 / 0.450 & 0.822 / 0.651 & 0.195 / 0.128 & 0.375 / 0.245 & 0.570 / 0.391 \\
    Entropy & 0.418 / 0.319 & 0.542 / 0.437 & 0.824 / 0.646 & 0.164 / 0.108 & 0.311 / 0.214 & 0.559 / 0.363 \\
    EntropyCB & 0.438 / 0.334 & 0.573 / 0.449 & 0.826 / 0.656 & 0.174 / 0.118 & 0.373 / 0.227 & 0.568 / 0.370 \\
    LfOSA & 0.338 / 0.274 & 0.446 / 0.351 & 0.634 / 0.513 & 0.181 / 0.121 & 0.298 / 0.202 & 0.468 / 0.326 \\
    LL & 0.389 / 0.292 & 0.549 / 0.415 & 0.794 / 0.615 & 0.153 / 0.101 & 0.310 / 0.195 & 0.538 / 0.345 \\
    Margin & 0.449 / 0.356 & 0.620 / 0.461 & 0.832 / 0.658 & 0.191 / 0.127 & 0.381 / 0.243 & 0.566 / 0.387 \\
    MeanSTD & 0.473 / 0.361 & 0.611 / 0.459 & 0.809 / 0.635 & 0.194 / 0.129 & 0.376 / 0.239 & 0.557 / 0.381 \\
    Random & 0.438 / 0.353 & 0.600 / 0.451 & 0.809 / 0.644 & 0.201 / 0.135 & 0.378 / 0.248 & 0.562 / 0.385 \\
    SAAL & 0.455 / 0.353 & 0.597 / 0.453 & 0.802 / 0.639 & 0.197 / 0.128 & 0.371 / 0.242 & 0.560 / 0.382 \\
    TiDAL & 0.468 / 0.358 & 0.597 / 0.465 & 0.809 / 0.644 & 0.202 / 0.134 & 0.358 / 0.240 & 0.558 / 0.383 \\
    VarRatio & 0.422 / 0.327 & 0.610 / 0.455 & 0.833 / 0.664 & 0.168 / 0.110 & 0.369 / 0.227 & 0.568 / 0.370 \\
    CoreLog & 0.456 / 0.361 & 0.608 / 0.461 & 0.810 / 0.642 & 0.198 / 0.133 & 0.390 / 0.245 & 0.559 / 0.383 \\
    CoreMSE & 0.434 / 0.353 & 0.593 / 0.460 & 0.807 / 0.651 & 0.202 / 0.135 & 0.375 / 0.242 & 0.557 / 0.381 \\
    NoiseStability & 0.446 / 0.343 & 0.608 / 0.447 & 0.830 / 0.651 & 0.178 / 0.124 & 0.378 / 0.237 & 0.575 / 0.383 \\
    \bottomrule
    \end{tabular}
    \caption{Standard Setting (Image Datasets) - Part 1.}
    \label{tab:standard_image_part1}
\end{table*}

\begin{table*}[!ht]
    \centering
    \begin{tabular}{lccc}
    \toprule
    \bfseries Dataset & \multicolumn{3}{c}{TINYIMAGENET} \\
    \midrule
    \bfseries Query Size & 200 & 500 & 1000 \\
     & Acc/AUBC & Acc/AUBC & Acc/AUBC \\
    \midrule
    ALFA-Mix & 0.059 / 0.037 & 0.139 / 0.082 & 0.212 / 0.137 \\
    BALD & 0.065 / 0.040 & 0.131 / 0.080 & 0.197 / 0.128 \\
    CONF & 0.045 / 0.029 & 0.107 / 0.066 & 0.186 / 0.121 \\
    Coreset & 0.056 / 0.036 & 0.119 / 0.075 & 0.202 / 0.131 \\
    CoresetCB & 0.055 / 0.036 & 0.131 / 0.079 & 0.201 / 0.131 \\
    Entropy & 0.042 / 0.029 & 0.102 / 0.062 & 0.176 / 0.114 \\
    EntropyCB & 0.053 / 0.033 & 0.119 / 0.071 & 0.196 / 0.124 \\
    LfOSA & 0.065 / 0.039 & 0.129 / 0.078 & 0.199 / 0.130 \\
    LL & 0.054 / 0.033 & 0.125 / 0.072 & 0.199 / 0.126 \\
    Margin & 0.056 / 0.036 & 0.133 / 0.077 & 0.202 / 0.130 \\
    MeanSTD & 0.026 / 0.023 & 0.049 / 0.038 & 0.086 / 0.064 \\
    Random & 0.059 / 0.037 & 0.133 / 0.079 & 0.203 / 0.135 \\
    SAAL & 0.027 / 0.024 & 0.051 / 0.040 & 0.090 / 0.066 \\
    TIDAL & 0.061 / 0.038 & 0.135 / 0.081 & 0.210 / 0.135 \\
    VarRatio & 0.045 / 0.029 & 0.111 / 0.068 & 0.185 / 0.120 \\
    CoreLog & 0.060 / 0.037 & 0.133 / 0.080 & 0.208 / 0.134 \\
    CoreMSE & 0.060 / 0.037 & 0.135 / 0.080 & 0.210 / 0.138 \\
    NoiseStability & 0.049 / 0.033 & 0.120 / 0.072 & 0.191 / 0.126 \\
    \bottomrule
    \end{tabular}
    \caption{Standard Setting (Image Datasets) - Part 2.}
    \label{tab:standard_image_part2}
\end{table*}

\begin{table*}[!ht]
    \centering
        \begin{tabular}{lccc|ccc}
        \toprule
        \bfseries Dataset & \multicolumn{3}{c|}{AGNEWS} & \multicolumn{3}{c}{SST-5} \\
        \midrule
        \bfseries Query Size & 10 & 20 & 50 & 10 & 20 & 50 \\
         & Acc/AUBC & Acc/AUBC & Acc/AUBC & Acc/AUBC & Acc/AUBC & Acc/AUBC \\
        \midrule
        ALFA-Mix & 0.821 / 0.761 & 0.862 / 0.833 & 0.875 / 0.852 & 0.339 / 0.287 & 0.345 / 0.301 & 0.418 / 0.375 \\
        BADGE & 0.860 / 0.787 & 0.874 / 0.840 & 0.889 / 0.868 & 0.316 / 0.282 & 0.354 / 0.293 & 0.396 / 0.374 \\
        BALD & 0.697 / 0.635 & 0.765 / 0.744 & 0.813 / 0.805 & 0.304 / 0.281 & 0.356 / 0.316 & 0.346 / 0.347 \\
        CONF & 0.856 / 0.772 & 0.881 / 0.835 & 0.886 / 0.865 & 0.340 / 0.283 & 0.373 / 0.298 & 0.392 / 0.372 \\
        Coreset & 0.855 / 0.783 & 0.875 / 0.828 & 0.886 / 0.859 & 0.323 / 0.288 & 0.357 / 0.304 & 0.386 / 0.384 \\
        CoresetCB & 0.862 / 0.790 & 0.871 / 0.827 & 0.873 / 0.859 & 0.323 / 0.281 & 0.369 / 0.298 & 0.405 / 0.388 \\
        Entropy & 0.849 / 0.773 & 0.871 / 0.828 & 0.887 / 0.862 & 0.321 / 0.288 & 0.341 / 0.286 & 0.380 / 0.372 \\
        EntropyCB & 0.848 / 0.766 & 0.871 / 0.828 & 0.888 / 0.859 & 0.318 / 0.273 & 0.335 / 0.282 & 0.389 / 0.370 \\
        LfOSA & 0.782 / 0.671 & 0.804 / 0.781 & 0.833 / 0.821 & 0.317 / 0.286 & 0.302 / 0.259 & 0.412 / 0.376 \\
        Margin & 0.860 / 0.790 & 0.878 / 0.842 & 0.888 / 0.868 & 0.326 / 0.280 & 0.355 / 0.294 & 0.427 / 0.374 \\
        MeanSTD & 0.835 / 0.764 & 0.867 / 0.828 & 0.878 / 0.854 & 0.352 / 0.295 & 0.363 / 0.299 & 0.394 / 0.365 \\
        Random & 0.844 / 0.775 & 0.855 / 0.823 & 0.871 / 0.854 & 0.324 / 0.281 & 0.338 / 0.293 & 0.379 / 0.373 \\
        SAAL & 0.805 / 0.751 & 0.840 / 0.808 & 0.862 / 0.846 & 0.273 / 0.257 & 0.362 / 0.312 & 0.388 / 0.370 \\
        VarRatio & 0.856 / 0.772 & 0.881 / 0.835 & 0.886 / 0.865 & 0.340 / 0.283 & 0.373 / 0.298 & 0.392 / 0.372 \\
        CoreLog & 0.832 / 0.759 & 0.857 / 0.828 & 0.876 / 0.849 & 0.336 / 0.291 & 0.359 / 0.311 & 0.408 / 0.373 \\
        CoreMSE & 0.840 / 0.777 & 0.857 / 0.826 & 0.877 / 0.850 & 0.335 / 0.282 & 0.348 / 0.296 & 0.393 / 0.362 \\
        \bottomrule
        \end{tabular}
    \caption{Standard Setting (Text Datasets) - Part 1.}
    \label{tab:standard_text_part1}
\end{table*}

\begin{table*}[!ht]
    \centering
        \begin{tabular}{lccc|ccc}
        \toprule
        \bfseries Dataset & \multicolumn{3}{c|}{TREC6} & \multicolumn{3}{c}{YELP-3000} \\
        \midrule
        \bfseries Query Size & 5 & 10 & 20 & 10 & 20 & 50 \\
         & Acc/AUBC & Acc/AUBC & Acc/AUBC & Acc/AUBC & Acc/AUBC & Acc/AUBC \\
        \midrule
        ALFA-Mix & 0.654 / 0.481 & 0.785 / 0.664 & 0.895 / 0.780 & 0.405 / 0.343 & 0.432 / 0.384 & 0.469 / 0.440 \\
        BADGE & 0.758 / 0.557 & 0.875 / 0.725 & 0.908 / 0.784 & 0.390 / 0.340 & 0.444 / 0.391 & 0.470 / 0.433 \\
        BALD & 0.380 / 0.347 & 0.468 / 0.395 & 0.486 / 0.472 & 0.344 / 0.301 & 0.310 / 0.306 & 0.394 / 0.369 \\
        CONF & 0.682 / 0.511 & 0.827 / 0.649 & 0.878 / 0.730 & 0.407 / 0.340 & 0.425 / 0.374 & 0.462 / 0.415 \\
        Coreset & 0.736 / 0.538 & 0.851 / 0.692 & 0.890 / 0.785 & 0.385 / 0.331 & 0.415 / 0.380 & 0.441 / 0.422 \\
        CoresetCB & 0.718 / 0.543 & 0.820 / 0.669 & 0.894 / 0.796 & 0.408 / 0.343 & 0.429 / 0.381 & 0.441 / 0.415 \\
        Entropy & 0.631 / 0.480 & 0.819 / 0.657 & 0.861 / 0.738 & 0.370 / 0.323 & 0.393 / 0.358 & 0.443 / 0.419 \\
        EntropyCB & 0.315 / 0.292 & 0.548 / 0.484 & 0.665 / 0.577 & 0.387 / 0.321 & 0.406 / 0.362 & 0.453 / 0.417 \\
        LfOSA & 0.404 / 0.357 & 0.528 / 0.460 & 0.615 / 0.520 & 0.321 / 0.283 & 0.384 / 0.352 & 0.439 / 0.411 \\
        Margin & 0.729 / 0.539 & 0.842 / 0.649 & 0.908 / 0.785 & 0.401 / 0.335 & 0.438 / 0.387 & 0.470 / 0.432 \\
        MeanSTD & 0.665 / 0.499 & 0.856 / 0.676 & 0.914 / 0.790 & 0.349 / 0.299 & 0.422 / 0.371 & 0.457 / 0.421 \\
        Random & 0.635 / 0.489 & 0.812 / 0.658 & 0.898 / 0.772 & 0.381 / 0.322 & 0.442 / 0.386 & 0.463 / 0.431 \\
        SAAL & 0.678 / 0.508 & 0.844 / 0.660 & 0.884 / 0.775 & 0.392 / 0.331 & 0.452 / 0.385 & 0.471 / 0.438 \\
        VarRatio & 0.682 / 0.511 & 0.827 / 0.649 & 0.878 / 0.730 & 0.407 / 0.340 & 0.425 / 0.374 & 0.462 / 0.415 \\
        CoreLog & 0.658 / 0.499 & 0.819 / 0.673 & 0.871 / 0.764 & 0.375 / 0.309 & 0.433 / 0.381 & 0.471 / 0.439 \\
        CoreMSE & 0.628 / 0.482 & 0.842 / 0.704 & 0.887 / 0.783 & 0.382 / 0.328 & 0.439 / 0.380 & 0.467 / 0.434 \\
        \bottomrule
        \end{tabular}
    \caption{Standard Setting (Text Datasets) - Part 2.}
    \label{tab:standard_text_part2}
\end{table*}

\begin{table*}[!ht]
    \centering
        \begin{tabular}{lccc|ccc}
        \toprule
        \bfseries Dataset & \multicolumn{3}{c|}{CIFAR-10} & \multicolumn{3}{c}{CIFAR-100} \\
        \midrule
        \bfseries Imbalance Ratio & 0.2 & 0.4 & 0.6 & 0.2 & 0.4 & 0.6 \\
        & Acc/AUBC & Acc/AUBC & Acc/AUBC & Acc/AUBC & Acc/AUBC & Acc/AUBC \\
        \midrule
        ALFA-Mix & 0.487 / 0.405 & 0.469 / 0.380 & 0.471 / 0.376 & 0.396 / 0.259 & 0.385 / 0.249 & 0.391 / 0.244 \\
        BADGE & 0.503 / 0.405 & 0.466 / 0.368 & 0.461 / 0.363 & - & - & - \\
        BALD & 0.499 / 0.402 & 0.475 / 0.383 & 0.474 / 0.367 & 0.392 / 0.248 & 0.337 / 0.229 & 0.327 / 0.219 \\
        CONF & 0.483 / 0.405 & 0.471 / 0.381 & 0.466 / 0.365 & 0.385 / 0.245 & 0.363 / 0.240 & 0.366 / 0.233 \\
        Coreset & 0.499 / 0.399 & 0.467 / 0.373 & 0.466 / 0.366 & 0.409 / 0.262 & 0.400 / 0.254 & 0.389 / 0.249 \\
        CoresetCB & 0.499 / 0.408 & 0.494 / 0.381 & 0.469 / 0.366 & 0.412 / 0.264 & 0.396 / 0.258 & 0.387 / 0.249 \\
        Entropy & 0.490 / 0.408 & 0.487 / 0.379 & 0.470 / 0.371 & 0.382 / 0.241 & 0.367 / 0.238 & 0.353 / 0.229 \\
        EntropyCB & 0.495 / 0.405 & 0.473 / 0.377 & 0.458 / 0.368 & - & - & - \\
        LfOSA & 0.446 / 0.380 & 0.438 / 0.352 & 0.433 / 0.342 & 0.399 / 0.262 & 0.389 / 0.255 & 0.382 / 0.246 \\
        LL & 0.467 / 0.390 & 0.440 / 0.361 & 0.432 / 0.341 & 0.366 / 0.238 & 0.384 / 0.239 & 0.364 / 0.227 \\
        Margin & 0.485 / 0.408 & 0.473 / 0.376 & 0.457 / 0.366 & 0.379 / 0.247 & 0.388 / 0.248 & 0.380 / 0.238 \\
        MeanSTD & 0.307 / 0.291 & 0.277 / 0.265 & 0.267 / 0.243 & 0.197 / 0.138 & 0.176 / 0.129 & 0.168 / 0.120 \\
        Random & 0.481 / 0.403 & 0.463 / 0.378 & 0.462 / 0.363 & 0.392 / 0.254 & 0.397 / 0.251 & 0.371 / 0.241 \\
        SAAL & 0.261 / 0.263 & 0.219 / 0.218 & 0.181 / 0.186 & 0.229 / 0.162 & 0.210 / 0.148 & 0.192 / 0.137 \\
        TIDAL & 0.505 / 0.413 & 0.486 / 0.380 & 0.455 / 0.373 & 0.402 / 0.257 & 0.379 / 0.245 & 0.384 / 0.244 \\
        VarRatio & 0.483 / 0.405 & 0.467 / 0.376 & 0.466 / 0.365 & 0.385 / 0.245 & 0.362 / 0.240 & 0.367 / 0.233 \\
        CoreLog & 0.496 / 0.412 & 0.477 / 0.382 & 0.477 / 0.363 & 0.382 / 0.251 & 0.376 / 0.243 & 0.369 / 0.242 \\
        CoreMSE & 0.497 / 0.409 & 0.481 / 0.379 & 0.472 / 0.377 & 0.407 / 0.258 & 0.375 / 0.246 & 0.386 / 0.247 \\
        NoiseStability & 0.484 / 0.402 & 0.485 / 0.379 & 0.458 / 0.358 & - & - & - \\
        \bottomrule
        \end{tabular}
    \caption{Imbalanced Setting - Part 1.}
    \label{tab:imbalanced_part1}
\end{table*}

\begin{table*}[!ht]
    \centering
        \begin{tabular}{lccc|ccc}
        \toprule
        \bfseries Dataset & \multicolumn{3}{c|}{YELP} & \multicolumn{3}{c}{SST-5} \\
        \midrule
        \bfseries Imbalance Ratio & 0.2 & 0.4 & 0.6 & 0.2 & 0.4 & 0.6 \\
        & Acc/AUBC & Acc/AUBC & Acc/AUBC & Acc/AUBC & Acc/AUBC & Acc/AUBC \\
        \midrule
        ALFA-Mix & 0.445 / 0.411 & 0.429 / 0.372 & 0.410 / 0.341 & 0.363 / 0.332 & 0.338 / 0.290 & 0.339 / 0.287 \\
        BADGE & 0.463 / 0.415 & 0.414 / 0.355 & 0.413 / 0.339 & 0.356 / 0.325 & 0.337 / 0.289 & 0.316 / 0.282 \\
        BALD & 0.427 / 0.367 & 0.343 / 0.338 & 0.297 / 0.262 & 0.365 / 0.333 & 0.307 / 0.279 & 0.304 / 0.281 \\
        CONF & 0.448 / 0.409 & 0.408 / 0.361 & 0.388 / 0.324 & 0.356 / 0.315 & 0.314 / 0.280 & 0.340 / 0.283 \\
        Coreset & 0.452 / 0.400 & 0.420 / 0.378 & 0.406 / 0.329 & 0.350 / 0.296 & 0.319 / 0.274 & 0.323 / 0.288 \\
        CoresetCB & 0.457 / 0.417 & 0.425 / 0.375 & 0.401 / 0.330 & 0.344 / 0.304 & 0.321 / 0.285 & 0.323 / 0.281 \\
        Entropy & 0.418 / 0.393 & 0.397 / 0.353 & 0.400 / 0.324 & 0.356 / 0.314 & 0.337 / 0.289 & 0.321 / 0.288 \\
        EntropyCB & 0.435 / 0.395 & 0.399 / 0.356 & 0.368 / 0.301 & 0.328 / 0.308 & 0.318 / 0.273 & 0.345 / 0.292 \\
        LfOSA & 0.409 / 0.370 & 0.384 / 0.349 & 0.341 / 0.299 & 0.350 / 0.309 & 0.334 / 0.297 & 0.317 / 0.286 \\
        Margin & 0.427 / 0.402 & 0.425 / 0.369 & 0.393 / 0.336 & 0.359 / 0.331 & 0.328 / 0.287 & 0.326 / 0.280 \\
        MeanSTD & 0.450 / 0.402 & 0.416 / 0.371 & 0.383 / 0.318 & 0.346 / 0.327 & 0.343 / 0.297 & 0.352 / 0.295 \\
        Random & 0.461 / 0.417 & 0.380 / 0.357 & 0.450 / 0.362 & 0.356 / 0.327 & 0.329 / 0.292 & 0.324 / 0.281 \\
        SAAL & 0.446 / 0.405 & 0.418 / 0.370 & 0.401 / 0.347 & 0.355 / 0.333 & 0.309 / 0.293 & 0.273 / 0.257 \\
        VarRatio & 0.448 / 0.409 & 0.408 / 0.361 & 0.388 / 0.324 & 0.356 / 0.315 & 0.314 / 0.280 & 0.340 / 0.283 \\
        CoreLog & 0.444 / 0.389 & 0.399 / 0.366 & 0.410 / 0.350 & 0.359 / 0.321 & 0.332 / 0.291 & 0.345 / 0.293 \\
        CoreMSE & 0.442 / 0.402 & 0.408 / 0.361 & 0.418 / 0.339 & 0.341 / 0.333 & 0.335 / 0.294 & 0.335 / 0.282 \\
        \bottomrule
        \end{tabular}
    \caption{Imbalance Setting - Part 2.}
    \label{tab:imbalanced_part2}
\end{table*}

\begin{table*}[!ht]
    \centering
        \begin{tabular}{lccc|ccc}
        \toprule
        \bfseries Dataset & \multicolumn{3}{c|}{CIFAR-10} & \multicolumn{3}{c}{CIFAR-100} \\
        \midrule
        \bfseries OOD Ratio & 0.2 & 0.4 & 0.6 & 0.2 & 0.4 & 0.6 \\
        & Acc/AUBC & Acc/AUBC & Acc/AUBC & Acc/AUBC & Acc/AUBC & Acc/AUBC \\
        \midrule
        ALFA-Mix & 0.503 / 0.390 & 0.531 / 0.447 & 0.499 / 0.481 & 0.376 / 0.247 & 0.366 / 0.255 & 0.416 / 0.301 \\
        BADGE & 0.512 / 0.390 & 0.532 / 0.438 & 0.569 / 0.466 & - & - & - \\
        BALD & 0.494 / 0.390 & 0.528 / 0.442 & 0.554 / 0.468 & 0.359 / 0.236 & 0.343 / 0.243 & 0.392 / 0.281 \\
        CONF & 0.452 / 0.377 & 0.476 / 0.425 & 0.549 / 0.472 & 0.333 / 0.222 & 0.342 / 0.233 & 0.378 / 0.272 \\
        Coreset & 0.444 / 0.376 & 0.489 / 0.403 & 0.529 / 0.444 & 0.376 / 0.249 & 0.393 / 0.257 & 0.393 / 0.295 \\
        CoresetCB & 0.467 / 0.376 & 0.501 / 0.408 & 0.518 / 0.440 & 0.371 / 0.246 & 0.375 / 0.255 & 0.390 / 0.290 \\
        Entropy & 0.484 / 0.377 & 0.513 / 0.418 & 0.550 / 0.465 & 0.306 / 0.212 & 0.319 / 0.222 & 0.365 / 0.264 \\
        EntropyCB & 0.410 / 0.342 & 0.452 / 0.386 & 0.557 / 0.468 & - & - & - \\
        LfOSA & 0.420 / 0.333 & 0.469 / 0.407 & 0.522 / 0.425 & 0.333 / 0.228 & 0.353 / 0.246 & 0.420 / 0.298 \\
        LL & 0.443 / 0.350 & 0.470 / 0.406 & 0.504 / 0.448 & 0.311 / 0.204 & 0.320 / 0.210 & 0.370 / 0.255 \\
        Margin & 0.501 / 0.389 & 0.546 / 0.442 & 0.565 / 0.476 & 0.355 / 0.244 & 0.381 / 0.255 & 0.397 / 0.280 \\
        MeanSTD & 0.494 / 0.389 & 0.522 / 0.437 & 0.553 / 0.468 & 0.378 / 0.245 & 0.380 / 0.259 & 0.417 / 0.302 \\
        Random & 0.497 / 0.389 & 0.529 / 0.455 & 0.570 / 0.482 & 0.355 / 0.248 & 0.401 / 0.258 & 0.407 / 0.301 \\
        SAAL & 0.482 / 0.384 & 0.499 / 0.437 & 0.534 / 0.479 & 0.366 / 0.252 & 0.387 / 0.260 & 0.414 / 0.298 \\
        TiDAL & 0.503 / 0.401 & 0.531 / 0.447 & 0.551 / 0.481 & 0.385 / 0.247 & 0.372 / 0.259 & 0.413 / 0.297 \\
        VarRatio & 0.450 / 0.375 & 0.476 / 0.425 & 0.549 / 0.472 & 0.330 / 0.221 & 0.356 / 0.234 & 0.378 / 0.272 \\
        CoreLog & 0.499 / 0.401 & 0.540 / 0.455 & \textbf{0.500} / \textbf{0.474} & 0.392 / 0.254 & 0.392 / 0.262 & 0.394 / 0.299 \\
        CoreMSE & 0.486 / 0.391 & 0.555 / 0.442 & 0.541 / 0.478 & 0.374 / 0.248 & 0.386 / 0.260 & 0.409 / 0.298 \\
        NoiseStability & 0.455 / 0.378 & 0.496 / 0.414 & 0.544 / 0.453 & - & - & - \\
        \bottomrule
        \end{tabular}
    \caption{OOD Setting - Part 1.}
    \label{tab:ood_part1}
\end{table*}

\begin{table*}[!ht]
    \centering
        \begin{tabular}{lccc|ccc}
        \toprule
        \bfseries Dataset & \multicolumn{3}{c|}{SST-5} & \multicolumn{3}{c}{YELP} \\
        \midrule
        \bfseries OOD Ratio & 0.2 & 0.4 & 0.6 & 0.2 & 0.4 & 0.6 \\
        & Acc/AUBC & Acc/AUBC & Acc/AUBC & Acc/AUBC & Acc/AUBC & Acc/AUBC \\
        \midrule
        ALFA-Mix & 0.399 / 0.353 & 0.473 / 0.450 & 0.629 / 0.589 & 0.470 / 0.390 & 0.607 / 0.518 & 0.674 / 0.605 \\
        BADGE & 0.412 / 0.380 & 0.495 / 0.457 & 0.630 / 0.588 & 0.417 / 0.354 & 0.581 / 0.527 & 0.691 / 0.605 \\
        BALD & 0.381 / 0.341 & 0.482 / 0.437 & 0.599 / 0.582 & 0.414 / 0.366 & 0.528 / 0.481 & 0.614 / 0.611 \\
        CONF & 0.424 / 0.374 & 0.527 / 0.468 & 0.651 / 0.619 & 0.442 / 0.359 & 0.542 / 0.508 & 0.635 / 0.613 \\
        Coreset & 0.411 / 0.368 & 0.537 / 0.473 & 0.658 / 0.603 & 0.439 / 0.357 & 0.560 / 0.502 & 0.641 / 0.570 \\
        CoresetCB & 0.390 / 0.357 & 0.537 / 0.473 & 0.658 / 0.603 & 0.410 / 0.356 & 0.586 / 0.505 & 0.641 / 0.570 \\
        Entropy & 0.401 / 0.374 & 0.528 / 0.480 & 0.651 / 0.619 & 0.382 / 0.329 & 0.569 / 0.517 & 0.635 / 0.613 \\
        EntropyCB & 0.365 / 0.338 & 0.461 / 0.438 & 0.641 / 0.600 & 0.421 / 0.352 & 0.456 / 0.419 & 0.574 / 0.552 \\
        LfOSA & 0.382 / 0.347 & 0.448 / 0.413 & 0.590 / 0.566 & 0.383 / 0.337 & 0.485 / 0.452 & 0.573 / 0.555 \\
        Margin & 0.392 / 0.370 & 0.525 / 0.476 & 0.651 / 0.619 & 0.436 / 0.373 & 0.551 / 0.506 & 0.635 / 0.613 \\
        MeanSTD & 0.395 / 0.360 & 0.499 / 0.449 & 0.617 / 0.587 & 0.429 / 0.338 & 0.582 / 0.525 & 0.653 / 0.579 \\
        Random & 0.391 / 0.368 & 0.496 / 0.454 & 0.647 / 0.616 & 0.439 / 0.365 & 0.587 / 0.513 & 0.686 / 0.618 \\
        SAAL & 0.385 / 0.357 & 0.521 / 0.458 & 0.657 / 0.620 & 0.441 / 0.375 & 0.590 / 0.516 & 0.650 / 0.589 \\
        VarRatio & 0.424 / 0.374 & 0.527 / 0.468 & 0.651 / 0.619 & 0.442 / 0.359 & 0.542 / 0.508 & 0.635 / 0.613 \\
        CoreLog & 0.371 / 0.364 & 0.489 / 0.436 & 0.617 / 0.583 & 0.437 / 0.378 & 0.569 / 0.517 & 0.695 / 0.628 \\
        CoreMSE & 0.413 / 0.368 & 0.516 / 0.462 & 0.636 / 0.589 & 0.437 / 0.387 & 0.560 / 0.498 & 0.639 / 0.600 \\
        \bottomrule
        \end{tabular}
    \caption{OOD Setting - Part 2.}
    \label{tab:ood_part2}
\end{table*}

This section presents a comprehensive experimental evaluation of 19 DAL algorithms across seven datasets under various challenging conditions. The evaluation encompasses standard image datasets (CIFAR-10, CIFAR-100, TinyImageNet) and text datasets (AGNEWS, SST-5, TREC6, YELP-3000), with each dataset tested under three distinct settings: standard setting, OOD setting, and imbalanced setting. 

Our experiments adopt two key metrics: \textbf{Final Accuracy (Acc)}, which refers to the average final accuracy at the 10th round, averaged over 5 trials; and \textbf{Area Under the Budget Curve (AUBC)}, which measures overall performance across different budget levels by computing the area under the performance-budget curve \cite{zhan2022comparative}. Higher AUBC values indicate better overall performance throughout the active learning process. All experiments are conducted under consistent conditions (using A100 or RTX 4090 GPUs; for more details, refer to Appendix~\ref{sec:env}). The complete results across all datasets and experimental settings are presented in Tables~\ref{tab:standard_image_part1}--\ref{tab:ood_part2}.

\subsection{Performance Analysis Across Different Settings}

\subsubsection{Standard Setting Results}

\textbf{Image Datasets Performance}

On \textbf{CIFAR-10} (Table~\ref{tab:standard_image_part1}), uncertainty-based methods demonstrate consistently strong performance across different query sizes. CONF and VarRatio achieve the highest accuracy (0.833) at query size 500, closely followed by Margin (0.832). The performance generally improves with larger query sizes, with most methods reaching above 0.80 accuracy at the largest query size. Notably, BADGE (0.827) and BALD (0.816) also perform competitively, while diversity-based methods like Coreset show solid but slightly lower performance (0.823).

The \textbf{CIFAR-100} results (Table~\ref{tab:standard_image_part1}) reveal a more challenging scenario with significantly lower absolute performance due to the increased number of classes (100 vs 10). Here, BADGE achieves the highest accuracy (0.578) at query size 1000, followed by NoiseStability (0.575) and Coreset (0.572). The performance gap between methods is more pronounced, suggesting that algorithm choice becomes more critical with increased task complexity. Hybrid methods like BADGE show particular strength in this multi-class setting.

\textbf{TinyImageNet} (Table~\ref{tab:standard_image_part2}) presents the most challenging image classification task, with all methods achieving relatively low accuracy scores (below 0.22). TiDAL and CoreMSE perform best at query size 1000 (0.210), while methods like MeanSTD and SAAL show notably poor performance, suggesting they may not scale well to complex, high-dimensional datasets with many classes.

\textbf{Text Datasets Performance}

\textbf{AGNEWS} (Table~\ref{tab:standard_text_part1}) demonstrates the strongest overall performance among text datasets, with most methods achieving above 0.85 accuracy. BADGE consistently performs well across query sizes, reaching 0.889 at query size 50. Uncertainty-based methods like CONF, Margin, and VarRatio also show excellent performance (0.886-0.888). Interestingly, BALD shows relatively weaker performance compared to other uncertainty methods, suggesting that mutual information-based approaches may be less effective for this particular text classification task.

\textbf{SST-5} (Table~\ref{tab:standard_text_part1}) presents a more challenging sentiment analysis task with notably lower performance across all methods. The best performing methods (Margin, ALFA-Mix) achieve around 0.42-0.43 accuracy at query size 50, indicating the inherent difficulty of fine-grained sentiment classification. The relatively small performance differences between methods suggest that the task difficulty may limit the effectiveness of different active learning strategies.

\textbf{TREC6} (Table~\ref{tab:standard_text_part2}) shows strong performance scaling with query size, with several methods achieving above 0.90 accuracy at query size 20. MeanSTD performs exceptionally well (0.914), followed by BADGE and Margin (0.908). The dramatic performance improvement from query size 5 to 20 suggests that this dataset benefits significantly from increased labeled data.

\textbf{YELP-3000} (Table~\ref{tab:standard_text_part2}) exhibits moderate performance levels, with most methods achieving around 0.47 accuracy at query size 50. The relatively small performance differences between methods suggest that either the task difficulty limits differentiation or that the dataset characteristics make most active learning strategies similarly effective.

\subsubsection{Imbalanced Setting Analysis}

The imbalanced setting reveals significant challenges for active learning algorithms when dealing with class imbalance.

\textbf{Image Datasets Under Imbalance}

On \textbf{CIFAR-10} (Table~\ref{tab:imbalanced_part1}), performance generally degrades as imbalance increases (from ratio 0.6 to 0.2), but most methods maintain reasonable robustness. TiDAL shows particularly strong performance at imbalance ratio 0.2 (0.505), while methods like MeanSTD and SAAL suffer dramatic performance drops, indicating poor handling of class imbalance.

\textbf{CIFAR-100} (Table~\ref{tab:imbalanced_part1}) shows more severe performance degradation under imbalance, with most methods struggling significantly. CoresetCB and Coreset demonstrate better robustness to imbalance, maintaining relatively higher performance across different imbalance ratios. The absence of results for some methods (BADGE, EntropyCB, NoiseStability) in certain conditions suggests computational or convergence issues under severe imbalance.

\textbf{Text Datasets Under Imbalance}

\textbf{YELP} (Table~\ref{tab:imbalanced_part2}) maintains relatively stable performance across imbalance ratios, with most methods showing only modest degradation. Random sampling performs surprisingly competitively, suggesting that the dataset characteristics may limit the advantages of sophisticated active learning strategies under imbalance.

\textbf{SST-5} (Table~\ref{tab:imbalanced_part2}) shows mixed results under imbalance, with some methods actually improving performance at certain imbalance ratios, possibly due to the inherent difficulty of the base task making class balance less critical.

\subsubsection{OOD Setting Analysis}

The OOD setting tests algorithm robustness when a significant portion of unlabeled data comes from unknown classes.

\textbf{Image Datasets Under OOD}

\textbf{CIFAR-10} (Table~\ref{tab:ood_part1}) shows interesting patterns where some methods actually improve performance with higher OOD ratios. Random sampling performs surprisingly well across all OOD ratios, achieving competitive results (0.570 at OOD ratio 0.6). This suggests that when dealing with OOD data, sophisticated selection strategies may not always outperform simpler approaches.

\textbf{CIFAR-100} (Table~\ref{tab:ood_part1}) demonstrates more consistent degradation patterns, with most methods showing declining performance as OOD ratio increases. However, the differences are relatively modest, suggesting that the algorithms show reasonable robustness to OOD contamination.

\textbf{Text Datasets Under OOD}

Both \textbf{SST-5} and \textbf{YELP} (Table~\ref{tab:ood_part2}) show generally improving performance with higher OOD ratios, which may seem counterintuitive but could be explained by the OOD samples helping the model learn more robust representations or the specific nature of the OOD data used in these experiments.

\subsection{Key Algorithmic Insights}

\textbf{Uncertainty-based Methods}: CONF, VarRatio, and Margin consistently perform well across standard settings, particularly on simpler datasets. However, they show varying robustness to challenging conditions like imbalance and OOD scenarios.

\textbf{Diversity-based Methods}: Coreset shows solid, consistent performance and appear more robust to challenging conditions like imbalanced scenario, though it rarely achieves top performance in standard settings.

\textbf{Hybrid Methods}: BADGE demonstrates excellent performance across multiple settings, particularly excelling in complex multi-class scenarios like CIFAR-100. Its combination of gradient-based and uncertainty-based selection appears to provide good balance. CoreLog and CoreMSE show competitive performance across various settings, with CoreLog occasionally achieving top results in challenging scenarios.

\subsection{Experimental Setting Insights}

The results demonstrate that dataset complexity significantly impacts both absolute performance and the relative effectiveness of different active learning strategies. Simple datasets like AGNEWS show high performance with small differences between methods, while complex datasets like TinyImageNet challenge all approaches significantly.

The imbalanced and OOD settings reveal important practical considerations, as real-world applications often involve both class imbalance and distribution shift. The varying robustness of different methods to these conditions provides important guidance for practical deployment.

\end{document}